\documentclass{article}
\usepackage{graphicx} 
\usepackage[a4paper, total={6in, 8in}]{geometry}
\usepackage{graphicx} 
\usepackage{amsmath,amssymb,amsfonts}%
\usepackage{amsthm}%
\usepackage{algorithmic}
\usepackage{algorithm}
\usepackage{graphics}
\usepackage{multirow} 
\usepackage{booktabs}

\usepackage{hyperref}
\usepackage{mathrsfs}%
\usepackage{authblk}

\title{Dynamic Angle Selection in X-Ray CT: A Reinforcement Learning Approach to Optimal Stopping}
\author[1,3]{Tianyuan Wang}
\author[1]{Felix Lucka}
\author[1]{Daniël M. Pelt}
\author[1,3]{K. Joost Batenburg}
\author[1,4]{Tristan van Leeuwen}

\affil[1]{Centrum Wiskunde \& Informatica, Science Park 123, Amsterdam, 1098 XG, The Netherlands}
\affil[2]{Leiden Institute of Advanced Computer Science, Leiden Universiteit, Leiden, The Netherlands}
\affil[3]{Mathematics Institute, Utrecht University, Campus-Boulevard 30, Utrecht, 3584 CD, The Netherlands}
\usepackage{xcolor}

\definecolor{myblue}{RGB}{28,130,185} 

\usepackage{comment}
\begin{document}

\maketitle

\section{Abstract}
In industrial X-ray Computed Tomography (CT), the need for rapid in-line inspection is critical. Sparse-angle tomography plays a significant role in this by reducing the required number of projections, thereby accelerating processing and conserving resources. Most existing methods aim to balance reconstruction quality and scanning time, typically relying on fixed scan durations. Adaptive adjustment of the number of angles is essential; for instance, more angles may be required for objects with complex geometries or noisier projections. The concept of optimal stopping, which dynamically adjusts this balance according to varying industrial needs, remains overlooked. Building on our previous work, we integrate optimal stopping into sequential Optimal Experimental Design (sOED) and Reinforcement Learning (RL). We propose a novel method for computing the policy gradient within the Actor-Critic framework, enabling the development of adaptive policies for informative angle selection and scan termination. Additionally, we investigate the gap between simulation and real-world applications in the context of the developed learning-based method. Our trained model, developed using synthetic data, demonstrates reliable performance when applied to experimental X-ray CT data. This approach enhances the flexibility of CT operations and expands the applicability of sparse-angle tomography in industrial settings. \\

\noindent\textbf{Keywords:} 
X-ray CT, optimal stopping, sequential optimal experimental design, policy gradient, reinforcement learning, simulation-to-real-world. 

\newpage

\section{Introduction}
X-ray Computed Tomography (CT) enables inline industrial inspection through three-dimensional reconstruction. However, fast and adaptive CT scanning is essential to make its widespread industrial application feasible. Previous studies have shown that not all projections are equally informative for certain objects \cite{kazantsev1991information, varga2011projection}. Most research has focused on sequentially selecting informative angles to enhance efficiency. This sequential Optimal Experimental Design (sOED) is often described within a Bayesian framework \cite{lindley1972bayesian,chaloner1995bayesian,rainforth2024modern}, where angles are chosen to maximize information gain. Information gain is typically quantified by comparing the prior and posterior distributions of the reconstruction or by assessing the similarity between the reconstructed image and the ground truth.

Batenburg \textit{et al.} \cite{batenburg2013dynamic} and Dabravolski \textit{et al.} \cite{dabravolski2014dynamic} used a set of template images composed of Gaussian blobs to represent samples from the prior distribution. They introduced an upper bound \cite{batenburg2011bounds} to approximate the information gain, which reflects the diameter of the solution set. Burger \textit{et al.} \cite{burger2021sequentially} employed classical “alphabetic criteria” for OED, such as "A-" and "D-" optimality, using the trace or determinant of the covariance matrix of the posterior distribution as summary statistics \cite{huan2024optimal}. Additionally, they used a Gaussian distribution as a prior, updating the posterior after selecting each angle. Building on this approach, Helin \textit{et al.} \cite{helin2022edge} introduced a Total Variation (TV) prior to enhance edges in reconstructions. The non-Gaussian TV prior was approximated as a Gaussian distribution using lagged diffusivity iteration. Furthermore, Barbano \textit{et al.} \cite{barbano2022bayesian} utilized a deep image prior as the reconstruction method and linearized the network to approximate the posterior distribution as a Gaussian. To avoid the need for a closed-form solution, Elata \textit{et al.} \cite{elata2025adaptive} proposed using a diffusion model for CT reconstruction, sampling from the posterior to approximate the posterior covariance matrix.

Recently, policy-based methods from the reinforcement learning community have been introduced into sOED \cite{shen2023bayesian,blau2022optimizing, foster2021deep}. In the medical CT field, Shen \textit{et al.} \cite{shen2022learning} trained a gated recurrent unit as a policy network on simulated medical data to map projections to probabilities over the angle space. We explored industrial CT applications with very few angles \cite{wang2024sequential}. We trained a policy that maps the current reconstruction to probabilities over the angle space. Additionally, we addressed a specific task—defect detection—by incorporating an extra reward for defect detectability and prior information about defects, enabling the trained policy to identify informative angles to aid in defect detection \cite{wang2024task}.

Although extensive work has been devoted to selecting informative projection angles, the question of how many angles to acquire is often neglected. This choice can be cast as an optimal stopping problem, a topic well studied in financial mathematics. The optimal expected reward is given by the Snell envelope, the smallest super‑martingale that dominates the reward process. The earliest (respectively, latest) optimal stopping time is the first instant at which the immediate reward equals (respectively, exceeds) the continuation value \cite{peskir2006optimal,becker2019deep,damera2024deep}.

Additionally, most studies have focused on simulated data rather than experimental X-ray CT data. This is particularly evident in learning-based methods, which rely on training with simulated data, leaving their generalizability to experimental X-ray CT data uncertain.

The contributions of this work include the development of an optimal stopping method to balance the trade-off between experimental costs and experimental goals, such as reconstruction quality. This approach enables both adaptive selection of informative angles and optimal scan termination based on experimental costs. By incorporating a terminal policy into the Actor-Critic framework, we proposed a novel method for computing the policy gradient, jointly optimizing angle selection and the terminal policy. Additionally, we investigated the gap between simulation and real-world applications by evaluating the trained model on experimental X-ray CT data.

The structure of the paper is as follows: The Background section provides an overview of the fundamental concepts and notations related to inverse problems and sOED using reinforcement learning. The Method section introduces our novel approach for computing the policy gradient. The Results section presents the findings from both simulation and experimental X-ray CT data experiments. Finally, the paper concludes with a discussion of the key findings and their implications.


\section{Background}

\subsection{Forward and Inverse Problems and Optimal Experimental Design (OED)}

Measurements \(\boldsymbol{y}(\boldsymbol{\theta})\) are obtained from the underlying parameters \(\Bar{\boldsymbol{x}}\) using a forward operator \(\boldsymbol{A}(\boldsymbol{\theta})\), which is determined by the design parameters \(\boldsymbol{\theta} = \left\{\theta_{1}, \ldots, \theta_{M} \right\}\). In the case of X-ray CT, the forward operator $\boldsymbol{A}(\boldsymbol{\theta})$ corresponds to the Radon transform for angles $\boldsymbol{\theta}$ \cite{hansen2021computed}. Since the measurement process is subject to noise, we also incorporate a noise term \(\boldsymbol{\epsilon}(\boldsymbol{\theta})\) and model the projections as follows:

\begin{equation}
    \boldsymbol{y}(\boldsymbol{\theta}) = \boldsymbol{A}(\boldsymbol{\theta})\Bar{\boldsymbol{x}} + \boldsymbol{\epsilon}(\boldsymbol{\theta}). 
    \label{eq:full_data_model}
\end{equation}

\begin{figure}[H]
  \centering
  \includegraphics[width=\linewidth]{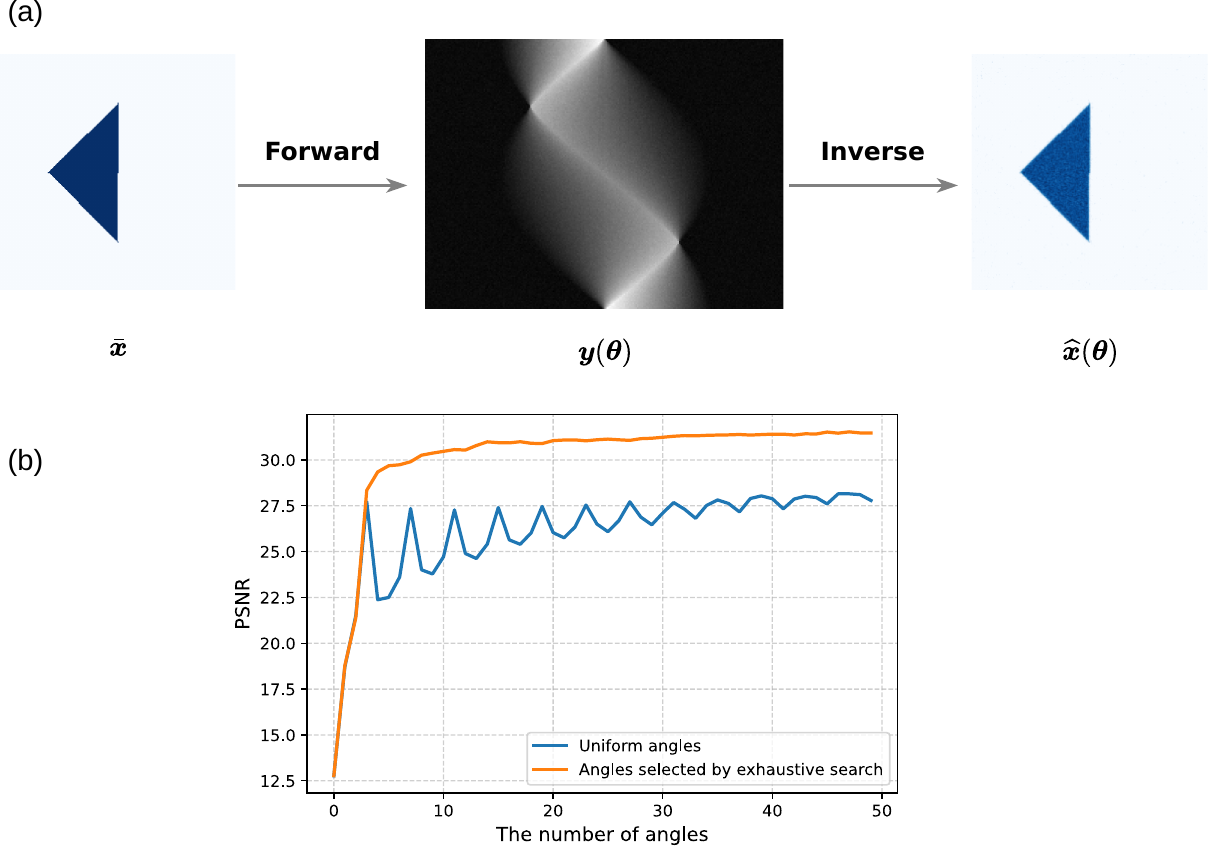}
  \caption{Triangle phantom example. (a) Forward process: noisy projections are generated by applying a random projection transform with $5\%$ Gaussian noise using parallel-beam geometry over 180 angles. Inverse process: the image is reconstructed from these noisy projections. (b) PSNR as a function of the number of angles for the triangle phantom. The number of angles increases in increments of one. The orange curve represents angles selected via exhaustive search, while the blue curve corresponds to uniformly spaced angles.
}
  \label{fig:intro}
\end{figure}

The inverse problem involves using the projections \(\boldsymbol{y}(\boldsymbol{\theta})\) and the forward operator \(\boldsymbol{A}(\boldsymbol{\theta})\) to reconstruct the underlying parameters \(\Bar{\boldsymbol{x}}\), denoted as \(\widehat{\boldsymbol{x}}(\boldsymbol{\theta})\). Figure (\ref{fig:intro}, a) illustrates the forward and inverse processes. However, solving inverse problems is often challenging due to their ill-posedness \cite{mueller2012linear}. The accuracy of the reconstructed underlying parameters is influenced by the design parameters \(\boldsymbol{\theta}\), making their optimal selection crucial. OED is employed to select the most informative angles to acquire the corresponding projections. To quantify the accuracy of the reconstruction, a utility function is defined within the OED framework. The optimal design \(\boldsymbol{\theta}^{*}\) is then obtained by maximizing the expected value of this utility function over the design space, taking into account the projections \(\boldsymbol{y}(\boldsymbol{\theta})\) and the underlying parameters \(\Bar{\boldsymbol{x}}\) \cite{lindley1972bayesian, ryan2016fully}. As shown in Figure (\ref{fig:intro}, b), we evaluated projection acquisition using 50 angles. The orange curve—which selects the most informative angles at each step via exhaustive search—consistently lies above the blue curve generated by uniformly spaced angles. Occasionally, the uniform strategy includes informative angles by chance, resulting in variability in its performance. Here, the Peak Signal-to-Noise Ratio (PSNR) serves as the utility function, quantifying reconstruction quality by comparing the reconstructed image to the ground truth. These results underscore the importance of optimal design when few angles are used. However, there is a clear trade-off: beyond a certain point, additional angles yield diminishing returns. For example, increasing the angle count from 10 to 50 produces only marginal PSNR gains. When projection acquisition is costly, such small improvements may not justify the added expense. Thus, identifying an optimal stopping point is essential to balance reconstruction quality against experimental costs.

\subsection{Sequential Optimal Experimental Design (sOED) and Reinforcement Learning}

Traditional experimental design is typically performed a-priori, with all optimal design parameters selected simultaneously. Because it ignores the feedback obtained after each parameter choice, this approach cannot support a-posteriori selection, potentially overlooking information that could refine subsequent decisions \cite{shen2023bayesian}.

The sOED extends the concept of traditional OED by allowing the design parameters to be determined sequentially, based on the data acquired from previous projections \cite{shen2023bayesian, rainforth2024modern}. For example, the design parameter \(\theta_k\) can be chosen based on the previous projections \(\boldsymbol{y}(\{\theta_{1}, \ldots, \theta_{k-1}\})\). This approach enables the design process to adapt dynamically to changes in the underlying parameters by adjusting the design parameters iteratively.

Solving OED problems in image reconstruction is inherently a bi-level optimization challenge: the lower level reconstructs the image, while the upper level minimizes the reconstruction error relative to the ground truth. The joint optimization problem is characterized by non-convexity, non-linearity, and high dimensionality, particularly in imaging applications \cite{ruthotto2018optimal}.
The sequential approach further complicates this optimization, as the bi-level optimization problem would need to be solved in real time. Reinforcement Learning (RL) \cite{sutton2018reinforcement}, a machine learning technique designed for fast sequential decision-making, facilitates the resolution of this sOED by training through interaction with the environment. RL is grounded in the framework of Markov Decision Processes (MDPs), which consist of a state space, action space, transition model, and reward function. In practice, MDPs are sometimes extended to Partially Observable MDPs (POMDPs) when the underlying state is not fully observable, necessitating the reconstruction of the belief state from measurements. The goal of RL is to learn a parameterized policy that maps the current state to the next action, maximizing the expectation of the cumulative rewards \cite{sutton2018reinforcement}. 

\begin{figure}[t]
  \centering
  \includegraphics[width=\linewidth]{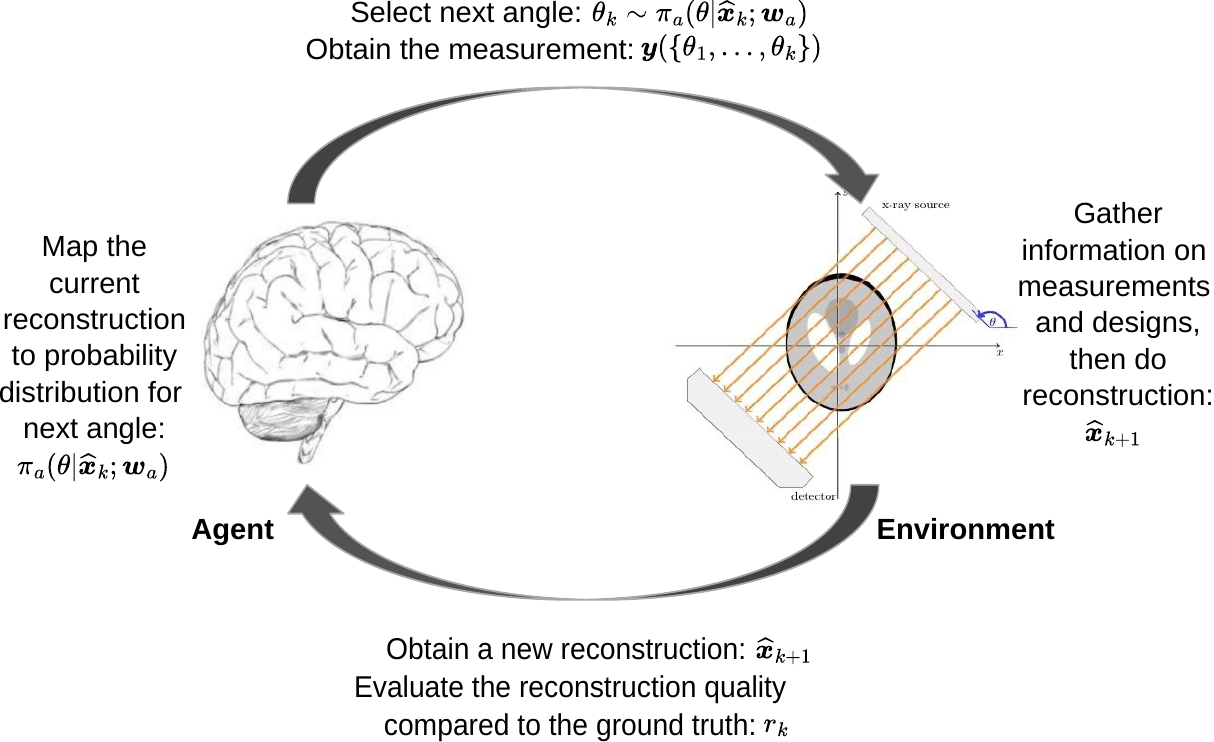}
  \caption{This figure illustrates how the reinforcement learning workflow maps onto the sOED framework. At the \(k\)th step, the action space comprises the possible values of the design parameter \(\theta_{k}\). Once \(\theta_{k}\) is chosen, the belief state is updated to the reconstructed estimate \(\widehat{\boldsymbol{x}}_{k+1}\), inferred from all prior projections \(\boldsymbol{y}(\{\theta_{1},\ldots,\theta_{k}\})\). The reward \(r_{k}\), representing reconstruction accuracy, is then computed.}
  \label{fig:Environment_agent}
\end{figure}

Figure (\ref{fig:Environment_agent}) illustrates how the RL framework integrates with the sOED process. In the context of sOED, at the \(k^{\text{th}}\) step,  the action space consists of the possible values that the design parameter \(\theta_{k}\) can take. After selecting \(\theta_{k}\), the reconstructed underlying parameters \(\widehat{\boldsymbol{x}}_{k+1}\), inferred from all previous projections \(\boldsymbol{y}(\{\theta_{1}, \ldots, \theta_{k}\})\), serve as the belief state. The utility function used for accuracy estimation acts as the reward function \(R\). For example, \(R\) can be defined as PSNR, which estimates the reconstruction quality by comparing the reconstruction with the ground truth \cite{wang2024sequential}. Consequently, the optimization problem of sOED is reformulated as learning a parameterized policy \(\pi_{\text{a}}(\theta_{k}|\widehat{\boldsymbol{x}}_{k};\boldsymbol{w}_{a})\), where \(\boldsymbol{w}_{a}\) represents the policy parameters. This policy, with the optimal parameters \(\boldsymbol{w}_{a}^{*}\), maximizes the expectation of the cumulative rewards obtained from the experiments.

By employing RL, the optimization process is simplified by focusing on optimizing the policy parameters rather than directly optimizing the design parameters. The state-value function \(V^{\pi_{\text{a}}}(\widehat{\boldsymbol{x}}_{1})\) represents the expected cumulative reward starting from the initial state \(\widehat{\boldsymbol{x}}_{1}\) and following the distribution \(\pi_{\text{chain}}\), which is defined as the product of the action selection policy \(\pi_{\text{a}}(\theta_{k}|\widehat{\boldsymbol{x}}_{k}; \boldsymbol{w}_{a})\) and the transition model \(\pi_{t}(\widehat{\boldsymbol{x}}_{k+1}|\widehat{\boldsymbol{x}}_{k}, \theta_{k})\). Formally, the state-value function is expressed as:

\begin{equation}
    V^{\pi_{\text{a}}}(\widehat{\boldsymbol{x}}_{1}) = \mathbb{E}_{\boldsymbol{\tau}\sim\pi_{\text{chain}}}\left[\sum_{k=1}^{M} r_{k} \mid \widehat{\boldsymbol{x}}_{1}\right],
\label{eq:ValueStandard}
\end{equation}

where the chain distribution is given by

\begin{equation}
    \pi_{\text{chain}}(\boldsymbol{\tau};\boldsymbol{w}_{a}) = \prod_{k=1}^{M} \pi_{\text{a}}(\theta_{k}|\widehat{\boldsymbol{x}}_{k}; \boldsymbol{w}_{a}) \, \pi_{t}(\widehat{\boldsymbol{x}}_{k+1}|\widehat{\boldsymbol{x}}_{k}, \theta_{k}).
\label{eq:ChainStandard}
\end{equation}

and a trajectory \(\boldsymbol{\tau}\) of \(M\) steps, \(\left\{\widehat{\boldsymbol{x}}_{1}, (\theta_{1}, \widehat{\boldsymbol{x}}_{2}, r_{1}), (\theta_{2}, \widehat{\boldsymbol{x}}_{3}, r_{2}), \ldots, (\theta_{M}, \widehat{\boldsymbol{x}}_{M+1}, r_{M})\right\}\), is sampled from \(\pi_{\text{chain}}\). Starting from the initial state \(\widehat{\boldsymbol{x}}_{1}\), each step involves selecting a design parameter \(\theta_{k}\), transitioning to the next state \(\widehat{\boldsymbol{x}}_{k+1}\), and accumulating the reward \(r_{k}\) associated with \(\widehat{\boldsymbol{x}}_{k+1}\).

The objective is to maximize the expected cumulative rewards across the trajectory.  For a fixed initial state \(\widehat{\boldsymbol{x}}_{1}\), the objective function of RL can be expressed in terms of the state-value function as:

\begin{equation}
    J(\boldsymbol{w}_{a}) = V^{\pi_{\text{a}}}(\widehat{\boldsymbol{x}}_{1}).
\label{eq:ObjectiveStandard}
\end{equation}

Two additional functions—the action-state value function and the advantage function—are introduced for gradient computation in the following section. The value function can be linked to actions through the action-state value function \(Q^{\pi_{\text{a}}}(\widehat{\boldsymbol{x}}, \theta)\), which estimates the expected cumulative rewards at the current state and specific action. The relationship between the state-value function and the action-value function is given by:

\begin{equation}
    V^{\pi_{\text{a}}}(\widehat{\boldsymbol{x}}) = \sum\limits_{\theta}\pi_{\text{a}}(\theta|\widehat{\boldsymbol{x}};\boldsymbol{w}_{a})Q^{\pi_{\text{a}}}(\widehat{\boldsymbol{x}}, \theta).
\label{eq:ReVQ}
\end{equation}

The advantage function quantifies how much better a specific action is compared to the average performance at a given state. It is defined as:

\begin{equation}
    A^{\pi_{\text{a}}}(\widehat{\boldsymbol{x}}, \theta) = Q^{\pi_{\text{a}}}(\widehat{\boldsymbol{x}}, \theta) - V^{\pi_{\text{a}}}(\widehat{\boldsymbol{x}}).
\end{equation}

\section{Method}
This section presents two approaches to optimal stopping. The first approach treats termination as an additional action in the action space. The second approach defines a separate termination policy, which is optimized jointly with the angle selection policy.

\subsection{Naive optimal stopping}
Optimal stopping for sOED can be implemented by introducing an additional terminal action within the action space. In our previous formulation \cite{wang2024sequential}, the action space comprised the 180 angles \(0^{\circ},\dots,179^{\circ}\), encoded as one‑hot vectors in \(\mathbb{R}^{180}\).  
In the present work, we extend this space by adding an explicit termination action; consequently, actions are encoded as one‑hot vectors vectors in \(\mathbb{R}^{181}\), where the first 180 basis vectors represent the angles and the \(181^{\text{st}}\) basis vector denotes termination. The \textit{reward function} for 'termination' and 'continuation' is defined by \( R(\widehat{\boldsymbol{x}}, \Bar{\boldsymbol{x}}, \theta) \), which accounts for the decision to either continue or terminate, as follows:

\[
R(\widehat{\boldsymbol{x}}, \Bar{\boldsymbol{x}}, \theta) =
\begin{cases}
-b, & \text{if } \theta = \theta_{max}, \\
\mathrm{PSNR}(\widehat{\boldsymbol{x}}, \Bar{\boldsymbol{x}}), & \text{if } \text{otherwise},
\end{cases}
\]

where \( \mathrm{PSNR}(\widehat{\boldsymbol{x}}, \Bar{\boldsymbol{x}}) \) is a function that serves as the immediate reward, evaluating the quality of the experiment at the stopping point \(\widehat{\boldsymbol{x}}\), and \( -b \) is a negative scaling factor representing the experimental cost incurred at each step. This mechanism works because the agent aims to maximize the cumulative reward: as long as selecting an additional angle leads to a \(\mathrm{PSNR}\) gain that exceeds the experimental cost for continuing, the agent will proceed. It will choose to stop only when further actions are expected to result in no meaningful improvement in PSNR relative to the experimental cost.

\begin{figure}[H]
  \centering
  \includegraphics[width=0.9\linewidth]{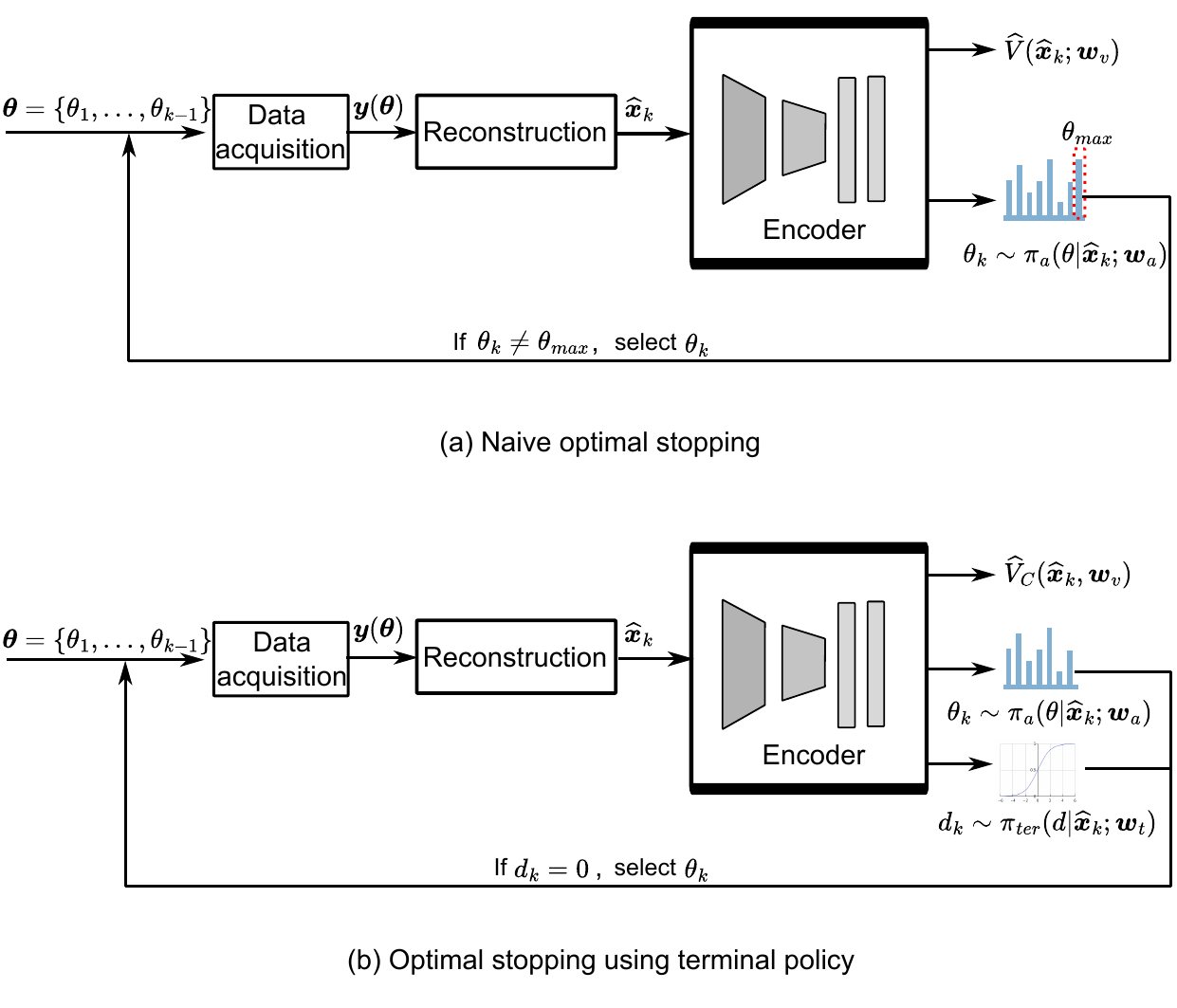}
  \caption{This figure illustrates the workflows: (a) Naive stopping: after selecting angles \(\{\theta_1,\dots,\theta_{k-1}\}\) and acquiring projections \(\boldsymbol{y}(\boldsymbol{\theta})\), the reconstruction \(\widehat{\boldsymbol{x}}_k\) becomes the belief state. A shared encoder then outputs (i) the state-value estimate \(\widehat V(\widehat{\boldsymbol{x}}_k;\boldsymbol{w}_v)\) and (ii) a distribution over all angles plus a terminal action \(\theta_{\max}\). (b) Terminal‐policy stopping: the encoder additionally outputs a probability distribution over termination versus continuation. Another distinction is that the top branch estimates the continuation value function using \(\widehat{V}_{C}(\widehat{\boldsymbol{x}}_{k}; \boldsymbol{w}_{v})\). In both workflows, a termination signal halts the process.
}
  \label{fig:all_algorithms}
\end{figure}

As described in our previous work \cite{wang2024sequential} and shown in Figure (\ref{fig:all_algorithms}, a), after selecting \(k-1\) angles \(\boldsymbol{\theta} = \left\{\theta_{1}, ..., \theta_{k-1}\right\}\), the \textit{observations} \(\boldsymbol{y}(\boldsymbol{\theta})\) are obtained from the data acquisition. The reconstructed image \(\widehat{\boldsymbol{x}}_{k}\) is then used as the \textit{belief state} and serves as the input to the shared encoder (initial belief state \(\widehat{\boldsymbol{x}}_{1}\) is a zero matrix), which has multi-head outputs. The shared encoder has two branches. One branch is designed to estimate the state-value function \( V^{\pi_{\text{a}}}(\widehat{\boldsymbol{x}})\), and we use \(\widehat{V}(\widehat{\boldsymbol{x}};\boldsymbol{w}_{v})\) for this approximation, where \(\boldsymbol{w}_{v}\) denotes the neural network parameters. The other branch outputs the distribution over the \textit{action space}, which consists of all possible angles and the terminal probability \(\theta_{\text{max}}\) as the final action. The inclusion of a mixture of actions for angles and termination represents the primary difference from the standard method described in our previous work \cite{wang2024sequential}. Consequently, the Temporal Difference (TD) error \cite{sutton2018reinforcement} for the terminal action is modified to account for the reward at the termination state \(\mathrm{PSNR}(\widehat{\boldsymbol{x}}_{k+1}, \Bar{\boldsymbol{x}})\), as shown in line 10 of Algorithm (\ref{alg:alg1}). Algorithm (\ref{alg:alg1}) outlines the modified algorithm based on the previous work \cite{wang2024sequential}.

\begin{algorithm}[H]
\caption{}\label{alg:alg1}
\begin{algorithmic}[1]
\STATE Initialize the action policy parameters $\boldsymbol{w}_{a}$, and the value function parameters $\boldsymbol{w}_{v}$ randomly. Define the maximal experimental steps $M$. Set step sizes $ \alpha^{\boldsymbol{w}_{a}} > 0$ and $\alpha^{\boldsymbol{w}_{v}} > 0$.
\STATE {\textbf{for each episode do:}}
\STATE \hspace{0.5cm} Get a phantom sample $\Bar{\boldsymbol{x}}$ then a zero matrix serves as the initial state \(\widehat{\boldsymbol{x}}_{1}\) and $k = 0$.
\STATE \hspace{0.5cm} {\textbf{while:}}
\STATE \hspace{1cm} Select the angle based on the Soft-max policy, which maps the inputs \\
\hspace{1cm} to a probability distribution that sums to 1: 
$\theta_{k} \sim \pi_{\text{a}} (\theta|\widehat{\boldsymbol{x}}_{k};\boldsymbol{w}_{a})$ \\
\STATE \hspace{1cm} Get new measurements $\mathbf{y}_{k}$ from the data acquisition \\
 \STATE \hspace{1cm} Reconstruct new image $\widehat{\boldsymbol{x}}_{k+1}$  \\
\STATE \hspace{1cm} Get reward for continuation \(-b\)
\\
\STATE  \hspace{1cm} Estimate the state-values $ \widehat{V}(\widehat{\boldsymbol{x}}_{k};\boldsymbol{w}_{v})$ and $ \widehat{V}(\widehat{\boldsymbol{x}}_{k+1};\boldsymbol{w}_{v})$ \\
\STATE \hspace{1cm} Compute TD error: \\
\hspace{1.1cm}{\textbf{if} $\theta_{k} = \theta_{\text{max}}$: \\
\hspace{1.5cm} Get reward for termination  \(\mathrm{PSNR}(\widehat{\boldsymbol{x}}_{k+1}, \Bar{\boldsymbol{x}})\) \\
\hspace{1.5cm}{ $\delta_{k} = -b +  \mathrm{PSNR}(\widehat{\boldsymbol{x}}_{k+1}, \Bar{\boldsymbol{x}})- \widehat{V}(\widehat{\boldsymbol{x}}_{k};\boldsymbol{w}_{v})$}} \\
\hspace{1.1cm}{\textbf{else:}} \\
    \hspace{1.5cm}{ $\delta_{k} = -b + \widehat{V}(\widehat{\boldsymbol{x}}_{k+1};\boldsymbol{w}_{v}) - \widehat{V}(\widehat{\boldsymbol{x}}_{k};\boldsymbol{w}_{v})$} \\
\STATE \hspace{1cm} Update the action policy function parameters $\boldsymbol{w}_{a}$:\\
\hspace{1cm} $\boldsymbol{w}_{a} \gets \boldsymbol{w}_{a} + \alpha^{\boldsymbol{w}_{a}}\nabla_{\boldsymbol{w}_{a}}\log \pi_{\text{a}}(\theta_{k}|\widehat{\boldsymbol{x}}_{k};\boldsymbol{w}_{a})\delta_{k} $
\STATE \hspace{1cm} Update value function parameters $\boldsymbol{w}_{v}$:\\
 \hspace{1cm} $\boldsymbol{w}_{v} \gets \boldsymbol{w}_{v} + \alpha^{\boldsymbol{w}_{v}}\nabla_{\boldsymbol{w}_{v}}\widehat{V}(\widehat{\boldsymbol{x}}_{k};\boldsymbol{w}_{v})\delta_{k}$
 \STATE \hspace{1cm} Increase the step number $k \mathrel{+}= 1$ \\
 \STATE \hspace{1cm} {\textbf{if} $\theta_{k} = \theta_{\text{max}}$ \textbf{or} $k = M$:} \\
 \hspace{1.5cm} \textbf{break}
\STATE \textbf{end for}
\end{algorithmic}
\label{alg1}
\end{algorithm}

\subsection{Optimal stopping using terminal policy}

\subsubsection{Objective function}
Different from the naive way of implementing optimal stopping, we consider an independent parameterized terminal policy \cite{bacon2017option}, which maps the state \( \widehat{\boldsymbol{x}} \) to a stochastic stopping decision \( d \sim \pi_{\text{ter}}(d|\widehat{\boldsymbol{x}}; \boldsymbol{w}_{t}) \). The continuation corresponds to \(d=0\), while termination corresponds to \(d=1\). Apart from the stopping decision, the reward function remains the same as in naive optimal stopping.




In the context of optimal stopping for sOED, the framework integrates both the terminal policy for scan termination and the action policy for angle selection, instead of combining them within the same action space as in the naive approach. First, we reformulate the probability chain from Equation (\ref{eq:ChainStandard}) to incorporate the termination decision. The updated probability chain reflects the decision-making process of the terminal policy, which evaluates at each step whether the trajectory should terminate or continue. 

Similar to Equation (\ref{eq:ChainStandard}), the probability chain \(\pi_{\text{chain,C}}(\boldsymbol{\tau}_{C}; \boldsymbol{w}_{a}, \boldsymbol{w}_{t})\) with the inclusion of the termination distribution is expressed as:

\begin{equation}
    \pi_{\text{chain,C}}(\boldsymbol{\tau}_{C}; \boldsymbol{w}_{a}, \boldsymbol{w}_{t}) = 
    \prod\limits_{k=1}^{T}\pi_{\text{sec}}(d_{k},\widehat{\boldsymbol{x}}_{k+1}, \theta_{k}|\widehat{\boldsymbol{x}}_{k};\boldsymbol{w}_{a}, \boldsymbol{w}_{t}),
\label{eq:TChain}        
\end{equation}

where the function \(\pi_{\text{sec}}\) is defined as:

\begin{equation}
    \pi_{\text{sec}}(d, \widehat{\boldsymbol{x}}^{'}, \theta|\widehat{\boldsymbol{x}}; \boldsymbol{w}_{a}, \boldsymbol{w}_{t}) = \pi_{\text{ter}}(d|\widehat{\boldsymbol{x}}; \boldsymbol{w}_{t}) +  \Bigl(1 - \pi_{\text{ter}}(d|\widehat{\boldsymbol{x}}; \boldsymbol{w}_{t}) \Bigr) \pi_{\text{a}}(\theta|\widehat{\boldsymbol{x}}; \boldsymbol{w}_{a}) \pi_{t}(\widehat{\boldsymbol{x}}^{'}|\widehat{\boldsymbol{x}}, \theta).
\label{eq:ChainTer}
\end{equation}

The stopping step \(T\) is determined by the terminal policy, which is less or equal to the maximal step \(M\) and \(d_{M} \equiv 1\). A trajectory up to the stopping point is generated from the new \(\pi_{\text{chain,C}}\) as:


\[
\left\{\widehat{\boldsymbol{x}}_{1}, (d_{1}, \theta_{1}, \widehat{\boldsymbol{x}}_{2}, -b), (d_{2}, \theta_{2}, \widehat{\boldsymbol{x}}_{3}, -b), \ldots, (d_{T}, \theta_{T}, \widehat{\boldsymbol{x}}_{T+1}, p_{T})\right\}.
\]

This trajectory represents the sequence of states (\(\widehat{\boldsymbol{x}}\)), termination indicators (\(d\)), experimental cost (\(-b\)), and quality evaluations \(\mathrm{PSNR}(\widehat{\boldsymbol{x}}_{T}, \Bar{\boldsymbol{x}})\) (\(p_{T}\)) at termination. The process is designed to terminate at step \(T\) as soon as \(d_{T} = 1\), ensuring that the trajectory is finite and explicitly concludes. For all prior steps (\(i < T\)), \(d_{i} = 0\), indicating that the process continues during those steps.

The objective function, representing the value function at the fixed initial state \(\widehat{\boldsymbol{x}}_{1}\), aims to learn both the action and terminal policies that maximize the expected cumulative rewards:
\begin{equation}
\begin{aligned}
J(\boldsymbol{w}_{a}, \boldsymbol{w}_{t}) &= V^{\pi_{\text{a}}, \pi_{\text{ter}}}(\widehat{\boldsymbol{x}}_{1}) \\
&= \pi_{\text{ter}}(d_{1}|\widehat{\boldsymbol{x}}_{1}; \boldsymbol{w}_{t})\,\mathrm{PSNR}(\widehat{\boldsymbol{x}}_{1}, \bar{\boldsymbol{x}}) \\
&\quad + \Bigl(1 - \pi_{\text{ter}}(d_{1}|\widehat{\boldsymbol{x}}_{1}; \boldsymbol{w}_{t})\Bigr) \, V_{C}^{\pi_{\text{a}}, \pi_{\text{ter}}}(\widehat{\boldsymbol{x}}_{1}).
\end{aligned}
\label{eq:OS objective}
\end{equation}

Here, the continuation state-value function is defined as follows \cite{damera2024deep, peskir2006optimal}:
\begin{equation}
V_{C}^{\pi_{\text{a}}, \pi_{\text{ter}}}(\widehat{\boldsymbol{x}}_{1}) = -b + \mathbb{E}_{\boldsymbol{\tau}_{C}^{(2)} \sim \pi_{\text{chain,C}}^{(2)}}\left[V^{\pi_{\text{a}}, \pi_{\text{ter}}}(\widehat{\boldsymbol{x}}_{2})\right],
\end{equation}
where \(\boldsymbol{\tau}_{C}^{(2)}\) represents the trajectory starting from \(\widehat{\boldsymbol{x}}_{2}\), following the probability chain described in Equation~(\ref{eq:TChain}).

Similarly, the continuation action-value function is defined only when the trajectory continues, as one of its inputs is the action: \(Q_{C}^{\pi_{\text{a}}, \pi_{\text{ter}}}(\widehat{\boldsymbol{x}}, \theta)\).

The optimal terminal policy \(\pi_{\text{ter}}^{*}(d_{k}|\widehat{\boldsymbol{x}}_{k}; \boldsymbol{w}_{t})\) at step \(k\) is defined as \cite{becker2019deep, peskir2006optimal, damera2024deep}:
\begin{equation}
\pi_{\text{ter}}^{*}(d_{k}|\widehat{\boldsymbol{x}}_{k}; \boldsymbol{w}_{t}) =
\begin{cases}
\mathbb{I}\Bigl(\mathrm{PSNR}(\widehat{\boldsymbol{x}}_{k}, \bar{\boldsymbol{x}}) \geq V_{C}^{\pi_{\text{a}}, \pi_{\text{ter}}}(\widehat{\boldsymbol{x}}_{k})\Bigr), & \text{if } k < M, \\
1, & \text{if } k = M,
\end{cases}
\end{equation}
where \(\mathbb{I}(\cdot)\) denotes the indicator function and \(M\) is the maximum number of steps in the experiment.


Figure (\ref{fig:all_algorithms}, b) illustrates the workflow of optimal stopping using the terminal policy. Compared to the naive optimal stopping, this approach includes an additional branch for the terminal policy, which uses the sample from Sigmoid function to determine continuation or termination. Another distinction is that the top branch estimates the continuation value function \(V^{\pi_{\text{a}}, \pi_{\text{ter}}}_{C}(\widehat{\boldsymbol{x}}_{k})\) using \(\widehat{V}_{C}(\widehat{\boldsymbol{x}}_{k}; \boldsymbol{w}_{v})\).

\subsubsection{Policy gradient}

To jointly solve the optimal stopping and action selection problems for sOED, we propose a novel policy gradient method. The policy gradients are calculated using Equation (\ref{eq:OS objective}).

For the gradient with respect to the action policy \(\boldsymbol{w}_{a}\), the detailed unrolling recursive derivation is provided in Appendix (\ref{appendix:A}). By sampling from \(N\) trajectories, the gradient is expressed as:


\begin{equation}
\begin{aligned}
   \nabla_{\boldsymbol{w}_{a}} J(\boldsymbol{w}_{a}, \boldsymbol{w}_{t})
    &= \nabla_{\boldsymbol{w}_{a}} V^{\pi_{\text{a}}, \pi_{\text{ter}}}(\widehat{\boldsymbol{x}}_{1}) \\
    & \propto\sum\limits_{n=1}^{N} \Biggl(\sum\limits_{k=1}^{T} \nabla_{\boldsymbol{w} _{a}}\log\pi_{\text{a}}(\theta_{k}|\widehat{\boldsymbol{x}}_{k};\boldsymbol{w}_{a})Q_{C}^{\pi_{\text{a}}, \pi_{\text{ter}}}(\widehat{\boldsymbol{x}}_{k}, \theta)\Biggr),
\end{aligned}
\label{eq:GradientW1S}
\end{equation}

To improve the stability and efficiency of action policy gradient computation, a baseline is introduced by replacing the continuation action-state value function in Equation (\ref{eq:GradientW1S}) with an advantage function \cite{sutton2018reinforcement}. The continuation advantage function is defined by incorporating the output of the terminal policy; further details are provided in the Appendix (\ref{appendix:A0}). The continuation advantage function is approximated as:

\begin{equation}
\begin{aligned}
     A_{C}^{\pi_{\text{a}}, \pi_{\text{ter}}}(\widehat{\boldsymbol{x}}, \theta) &\approx  -b +  \Bigl(1-\pi_{\text{ter}}(d^{'}|\widehat{\boldsymbol{x}}^{'};\boldsymbol{w}_{t}) \Bigr)V_{C}^{\pi_{\text{a}}, \pi_{\text{ter}}}(\widehat{\boldsymbol{x}}^{'}) \\
     &+ \pi_{\text{ter}}(d^{'}|\widehat{\boldsymbol{x}}^{'};\boldsymbol{w}_{t}) \mathrm{PSNR}(\widehat{\boldsymbol{x}}^{'}, \Bar{\boldsymbol{x}}) - V_{C}^{\pi_{\text{a}}, \pi_{\text{ter}}}(\widehat{\boldsymbol{x}}).
\end{aligned}
\label{eq:ApAdvantage}
\end{equation}

We perform stochastic gradient ascent to update the parameters of the angle policy, \(\boldsymbol{w}_{a}\), using a step size of \(\alpha^{\boldsymbol{w}_{a}}\).

\begin{equation}
    \boldsymbol{w}_{a} \gets \boldsymbol{w}_{a} + \alpha^{\boldsymbol{w}_{a}} \nabla_{\boldsymbol{w}_{a}}\log\pi_{\text{a}}(\theta|\widehat{\boldsymbol{x}};\boldsymbol{w}_{a})A_{C}^{\pi_{\text{a}}, \pi_{\text{ter}}}(\widehat{\boldsymbol{x}},\theta)
\end{equation}

For the gradient with respect to the terminal policy \(\boldsymbol{w}_{t}\), the detailed unrolling recursive derivation is provided in the Appendix (\ref{appendix:B}). By sampling from \(N\) trajectories, the gradient is expressed as:


\begin{equation}
\begin{aligned}
   \nabla_{\boldsymbol{w}_{t}} J(\boldsymbol{w}_{a}, \boldsymbol{w}_{t})
    &= \nabla_{\boldsymbol{w}_{t}} V^{\pi_{\text{a}}, \pi_{\text{ter}}}(\widehat{\boldsymbol{x}}) \\
    & \propto \sum\limits_{n=1}^{N} \Biggl(\sum\limits_{k=1}^{T} \nabla_{\boldsymbol{w}_{t}}\pi_{\text{ter}}(d_{k}|\widehat{\boldsymbol{x}}_{k};\boldsymbol{w}_{t}) \Bigl(\mathrm{PSNR}(\widehat{\boldsymbol{x}}_{k}, \Bar{\boldsymbol{x}})-V_{C}^{\pi_{\text{a}}, \pi_{\text{ter}}}(\widehat{\boldsymbol{x}}_{k}) \Bigr)\Biggr),
\end{aligned}
\label{eq:GradientW2S}
\end{equation}

We also perform stochastic gradient ascent to update the parameters of the terminal policy, \(\boldsymbol{w}_{t}\), using a step size of \(\alpha^{\boldsymbol{w}_{t}}\).
\begin{equation}
    \boldsymbol{w}_{t} \gets \boldsymbol{w}_{t} + \alpha^{\boldsymbol{w}_{t}} \nabla_{\boldsymbol{w}_{t}}\pi_{\text{ter}}(d|\widehat{\boldsymbol{x}};\boldsymbol{w}_{t}) \Bigl(\mathrm{PSNR}(\widehat{\boldsymbol{x}}, \Bar{\boldsymbol{x}})-V_{C}^{\pi_{\text{a}}, \pi_{\text{ter}}}(\widehat{\boldsymbol{x}})) \Bigr)
\end{equation}

The complete algorithm is presented in Algorithm (\ref{alg:alg2}).


\begin{algorithm}[H]
\caption{}\label{alg:alg2}
\begin{algorithmic}[1]
\STATE Initialize the action policy parameters $\boldsymbol{w}_{a}$, the terminal policy parameters $\boldsymbol{w}_{t}$, and the value function parameters $\boldsymbol{w}_{v}$ randomly. Define the maximal experimental steps $M$. Set step sizes $ \alpha^{\boldsymbol{w}_{a}} > 0$, $\alpha^{\boldsymbol{w}_{t}} > 0$, $\alpha^{\boldsymbol{w}_{v}} > 0$.
\STATE {\textbf{for each episode do:}}
\STATE \hspace{0.5cm} Get a phantom sample $\Bar{\boldsymbol{x}}$ then a zero matrix serves as the initial state \(\widehat{\boldsymbol{x}}_{1}\) and $k = 0$.
\STATE \hspace{0.5cm} {\textbf{while $d_{k} = 0$ and $k < M$:}}
\STATE \hspace{1cm} Select the angle based on the Soft-max policy, which maps the inputs \\
\hspace{1cm} to a probability distribution that sums to 1: 
$\theta_{k} \sim \pi_{\text{a}} (\theta|\widehat{\boldsymbol{x}}_{k};\boldsymbol{w}_{a})$
\STATE \hspace{1cm} Get new measurements $\mathbf{y}_{k}$ from the data acquisition \\
 \STATE \hspace{1cm} Reconstruct new image $\widehat{\boldsymbol{x}}_{k+1}$  \\
\STATE \hspace{1cm} Get reward for continuation \(-b\), and rewards for termination \(\mathrm{PSNR}(\widehat{\boldsymbol{x}}_{k}, \Bar{\boldsymbol{x}})\) \\
\hspace{1cm} and \(\mathrm{PSNR}(\widehat{\boldsymbol{x}}_{k+1}, \Bar{\boldsymbol{x}})\)\\
\STATE \hspace{1cm} {Determine the terminal action based on the Sigmoid policy, which \\
\hspace{1cm} maps the inputs to terminal probability: $d_{k} \sim \pi_{\text{ter}}(d|\widehat{\boldsymbol{x}}_{k}; \boldsymbol{w}_{t})$}
\STATE  \hspace{1cm} Estimate the state-values $ \widehat{V}_{C}(\widehat{\boldsymbol{x}}_{k};\boldsymbol{w}_{v})$ and $ \widehat{V}_{C}(\widehat{\boldsymbol{x}}_{k+1}; \boldsymbol{w}_{v})$ \\
\STATE \hspace{1cm} {Compute TD error:  $\delta_{k} = -b +  \Bigl(1-\pi_{\text{ter}}(d_{k+1}|\widehat{\boldsymbol{x}}_{k+1};\boldsymbol{w}_{t}) \Bigr) \widehat{V}_{C}(\widehat{\boldsymbol{x}}_{k+1}; \boldsymbol{w}_{v})$\\
 \hspace{1cm}  + $\pi_{\text{ter}}(d_{k+1}|\widehat{\boldsymbol{x}}_{k+1};\boldsymbol{w}_{t}) \mathrm{PSNR}(\widehat{\boldsymbol{x}}_{k+1}, \Bar{\boldsymbol{x}})- \widehat{V}_{C}(\widehat{\boldsymbol{x}}_{k};\boldsymbol{w}_{v})$} \\
\STATE \hspace{1cm} Update policy function parameters $\boldsymbol{w}_{a}$:\\
\hspace{1cm} $\boldsymbol{w}_{a} \gets \boldsymbol{w}_{a} + \alpha^{\boldsymbol{w}_{a}}\nabla_{\boldsymbol{w}_{a}}\log \pi_{\text{a}}(\theta_{k}|\widehat{\boldsymbol{x}}_{k};\boldsymbol{w}_{a})\delta_{k} $
\STATE \hspace{1cm} {Update policy function parameters $\boldsymbol{w}_{t}$:\\
\hspace{1cm} $\boldsymbol{w}_{t} \gets \boldsymbol{w}_{t} + \alpha^{\boldsymbol{w}_{t}}\nabla_{\boldsymbol{w}_{t}}\pi_{\text{ter}}(d_{k}|\widehat{\boldsymbol{x}}_{k};\boldsymbol{w}_{t}) \Bigl(\mathrm{PSNR}(\widehat{\boldsymbol{x}}_{k}, \Bar{\boldsymbol{x}}) - \widehat{V}_{C}(\widehat{\boldsymbol{x}}_{k};\boldsymbol{w}_{v}) \Bigr) $}
\STATE \hspace{1cm} Update value function parameters $\boldsymbol{w}_{v}$:\\
 \hspace{1cm} $\boldsymbol{w}_{v} \gets \boldsymbol{w}_{v} + \alpha^{\boldsymbol{w}_{v}}\nabla_{\boldsymbol{w}_{v}}\widehat{V}_{C}(\widehat{\boldsymbol{x}}_{k};\boldsymbol{w}_{v})\delta_{k}$
 \STATE \hspace{1cm} Increase the step number $k \mathrel{+}= 1$ \\
\STATE \textbf{end for}
\end{algorithmic}
\label{alg2}
\end{algorithm}


\section{Results}

The proposed optimal stopping method for sOED was validated through a series of X-ray CT experiments. The model was first trained using synthetic data and then evaluated on experimental X-ray CT data to test its performance and generalizability. 

\subsection{Dataset}
\subsubsection{Synthetic dataset}
\begin{figure}[t]
  \centering
  \includegraphics[width=0.5\linewidth]{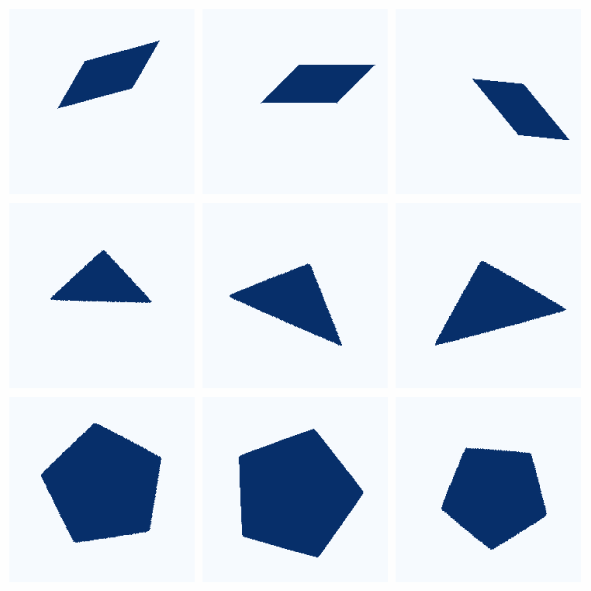}
  \caption{The figure shows nine samples from the synthetic dataset, including parallelograms, triangles, and pentagons. Each shape type is represented by three samples.}
  \label{fig:DatasetTraining}
\end{figure}

We created a synthetic dataset that includes features with strongly non-uniform orientations, corresponding to highly informative imaging angles. The dataset includes three shapes—parallelograms, triangles, and pentagons—each varying in scale, rotation, and position. These variations result in differing requirements for the number of angles needed to achieve accurate reconstructions. Figure (\ref{fig:DatasetTraining}) provides example images from this dataset, while the parameters used for its generation are outlined in the Appendix (\ref{appendix:D}).

\subsubsection{Experimental dataset}
\begin{figure}[t]
  \centering
  \includegraphics[width=0.8\linewidth]{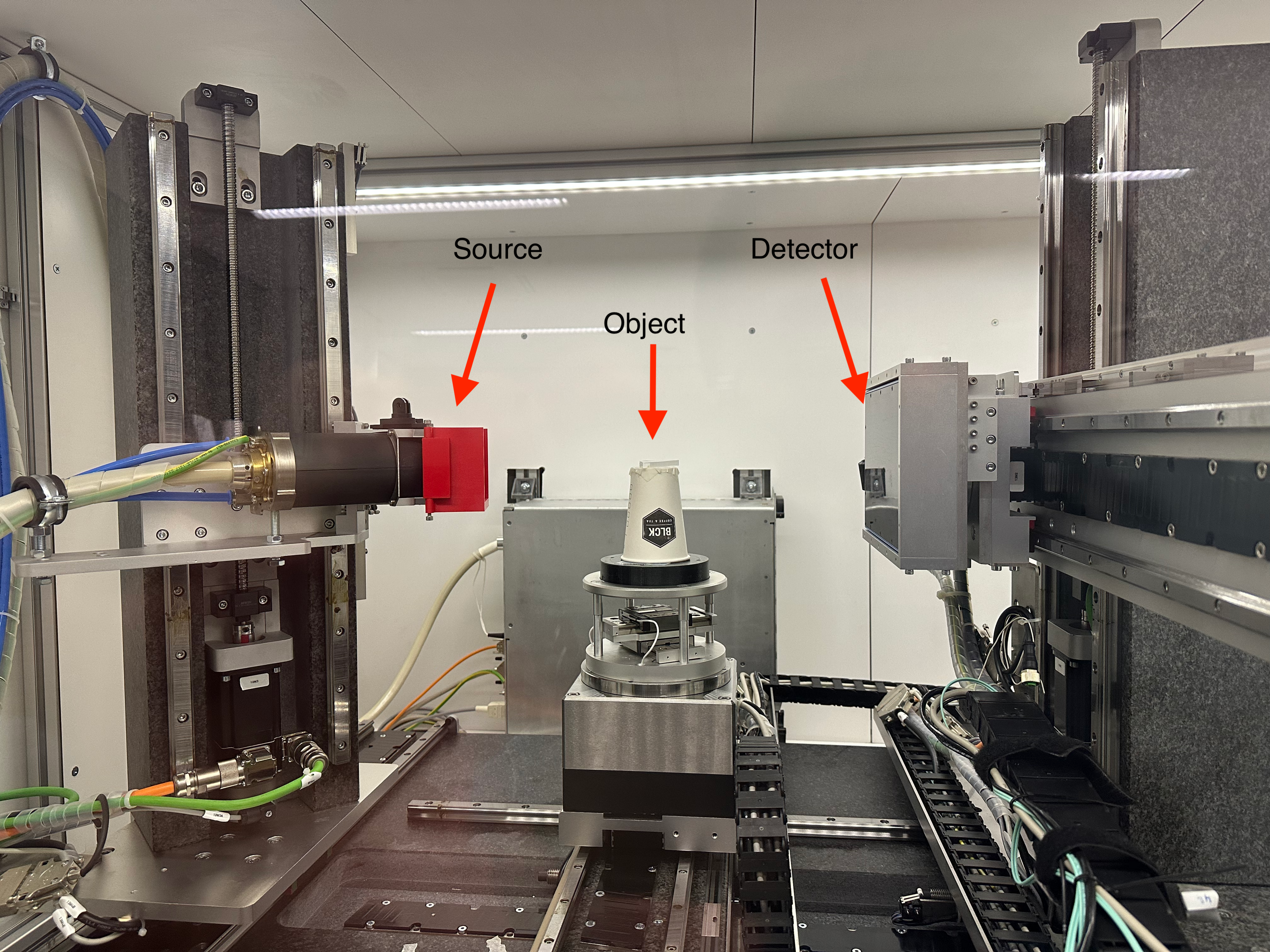}
  \caption{Experimental scanning setup at the FleX‑ray laboratory. The X‑ray source (left) and flat‑panel detector (right) remain fixed, while the object is positioned on a paper cup atop the rotation stage centered between them. During acquisition, the stage rotates to capture projections from multiple angles.}
  \label{fig:scanning_setting}
\end{figure}

All data were collected at the FleX-ray laboratory of Centrum Wiskunde en Informatica (CWI) in Amsterdam, the Netherlands \cite{wang_2025_14893740}. A scanning approach was used to acquire projections with dimensions of 956 \(\times\) 10 pixels. The source-to-object and detector-to-object distances were both set to 225 mm. An exposure time of 80 ms per projection was applied, and the source spectrum was shaped using filters consisting of 0.1 mm zinc, 0.2 mm copper, and 0.5 mm aluminum. A total of 3601 projections were acquired. Figure (\ref{fig:scanning_setting}) illustrates the scanning setup.

The dataset comprises two laser-cut objects—triangular and pentagonal—fabricated from 6 mm-thick transparent acrylate. Each shape is represented by 12 samples of varying sizes. The right-angle edges of the triangular samples range from 2.8 cm to 4.0 cm, while the edges of the pentagonal samples range from 2.5 cm to 3.0 cm. Each scanning session involved different placements of the objects, resulting in variations in rotation and translation. To create a dataset with two noise levels, two emission currents were used: 600 \(\mu\)A and 100 \(\mu\)A. The lower emission current (100 \(\mu\)A) produced data with higher noise levels. Finally, 12 groups of projections were acquired for each shape and each noise level, resulting in a total of 48 groups of projections.

For the purpose of this study, the number of projections was reduced to 361 by subsampling every 10th projection. To approximate the cone-beam geometry as a fan-beam geometry, only the middle row of the detector was used. The column size of the projections was then reduced to 239 by selecting every fourth pixel on the detector. Furthermore, the fan-beam data was rebinned into a parallel-beam data with 180 projections to simplify angle selection. This transformation made it more precise to identify the informative angles, as they are tangential to the edges in parallel-beam geometry.

\subsection{Implementation}
For training on the synthetic data, the Astra Toolbox \cite{van2015astra, van2016fast} was used to generate the simulated projections. For the image reconstruction \(\widehat{\boldsymbol{x}}(\boldsymbol{\theta})\), the SIRT algorithm with a non-negativity constraint was applied for 150 iterations.

For the algorithm implementation, the architectures of the encoder and the Actor-Critic neural networks are detailed in the Appendix (\ref{appendix:E}). During training, weights of 1.0 and 0.5 were assigned to the actor loss and critic loss in Algorithm (\ref{alg:alg1}) and Algorithm (\ref{alg:alg2}), respectively. To encourage exploration during training, an entropy loss with a weight of 0.01 was included. Compared to Algorithm (\ref{alg:alg1}), Algorithm (\ref{alg:alg2}) included an additional loss term for the terminal policy with a weight of 1.0. These parameter settings were empirically chosen to strike an optimal balance between policy optimization, accurate value estimation, and robust exploration during training. The network weights were optimized using the Adam optimizer \cite{kingma2014adam} with a learning rate of $10^{-4}$ and a weight decay of $10^{-5}$. To prevent selecting the same angle multiple times, previously chosen angles are masked according to the procedure described in \cite{huang2020closer}.

\subsection{Training on synthetic data with Gaussian noise}
In this experiment, we investigated whether the two algorithms could adaptively determine the optimal number of angles for CT imaging by adjusting the experimental costs before termination. The projections incorporated 5\% Gaussian noise. For Algorithms (\ref{alg:alg1}) and (\ref{alg:alg2}), the experimental costs ranged from -0.4 to -0.9 in intervals of 0.1. At higher experimental costs, priority is placed on guaranteeing image quality, whereas at lower costs the emphasis shifts toward reducing the number of projections. A separate model was trained for each experimental cost setting using the synthetic dataset shown in Figure (\ref{fig:DatasetTraining}). To prevent excessively non-termination during training, the maximum number of angles, \(M\), in Algorithm (\ref{alg:alg1}) and Algorithm (\ref{alg:alg2}), was limited to 20.

\begin{figure}[H]
  \centering
  \includegraphics[width=\linewidth]{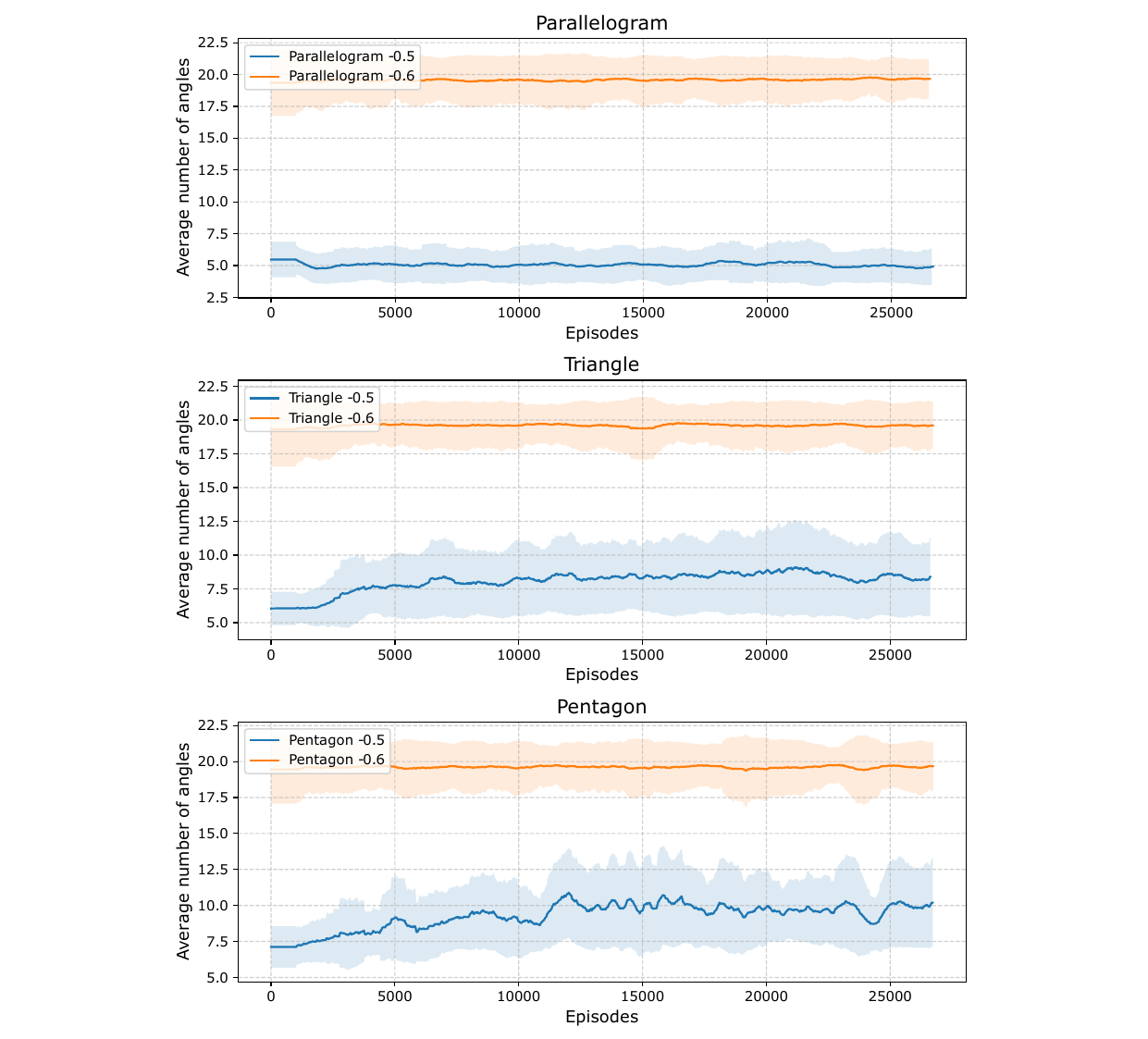}
  \caption{Comparison of the number of angles selected by the naive policy and by experimental cost settings of \(-0.5\) and \(-0.6\). The curves represent the mean number of angles during training, averaged over every 1000 episodes, and the shaded regions indicate the variance. Results are grouped by shape: parallelogram, triangle, and pentagon.}

\label{fig:Training_comparison_naive}
\end{figure}

During training, each episode involved sampling a data point from the dataset shown in Figure (\ref{fig:DatasetTraining}). The policy was trained at each angle selection step until termination, thereby completing the episode. A total of 80{,}000 episodes (i.e., 80{,}000 sampled data) were considered. However, the naive stopping mechanism from Algorithm (\ref{alg:alg1}) failed to function as intended. For most reward values ranging from -0.4 to -0.9, the algorithm consistently reached the maximum number of angles. As shown in Figure (\ref{fig:Training_comparison_naive}), the experimental cost setting of \(-0.5\) yielded different numbers of angles for each shape, whereas the \(-0.6\) setting failed to produce valid results. This outcome indicates that the policy gradient lacked robustness, failed to converge reliably in the intended direction, and tended to ignore the terminal action.

\begin{figure}[H]
  \centering
  \includegraphics[width=\linewidth]{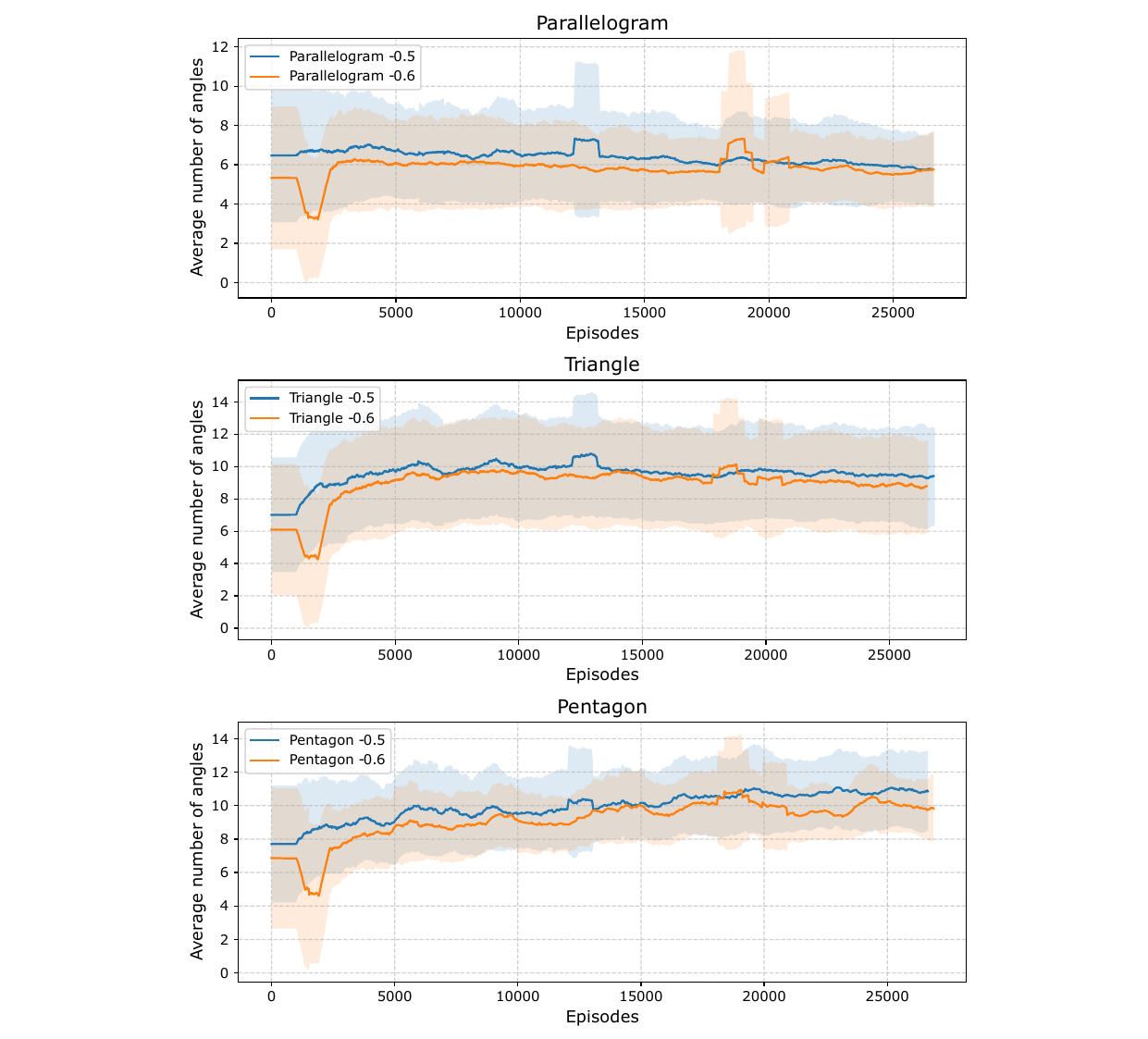}
  \caption{Comparison of the number of angles selected by the terminal policy and by experimental cost settings of \(-0.5\) and \(-0.6\). The curves represent the mean number of angles during training, averaged over every 1000 episodes, and the shaded regions indicate the variance. Results are grouped by shape: parallelogram, triangle, and pentagon.}

\label{fig:Training_comparison_terminal}
\end{figure}

In contrast, the terminal policy remained robust under different reward settings. Figure (\ref{fig:Training_comparison_terminal}) shows that, during training, the number of angles increased, and more complex shapes required more angles. Lower costs (experimental cost is -0.5) generally led to a higher number of angles.

\subsection{Validation on unseen rotations and noise levels}

To assess the generalizability of Algorithm (\ref{alg:alg2}) for optimal stopping, we applied the policy trained in the previous experiment (which involved \(5\%\) Gaussian noise) to synthetic data featuring unseen rotations—rotations not included in the training set shown in Figure (\ref{fig:DatasetTraining})—as well as two additional noise levels: \(3\%\) and \(7\%\) Gaussian noise. The total number of unseen phantoms was 1{,}800, with 600 for each shape.

A standard baseline approach, the \emph{Golden Ratio (GR) Policy} \cite{kohler2004projection,craig2023real}, is considered for comparison with the RL policy. In the GR policy, an angular increment based on an irrational number is used, which leads to a non-repeating sequence of angles that fill the angular space most evenly over time.


\begin{figure}[H]
  \centering
  \includegraphics[width=\linewidth]{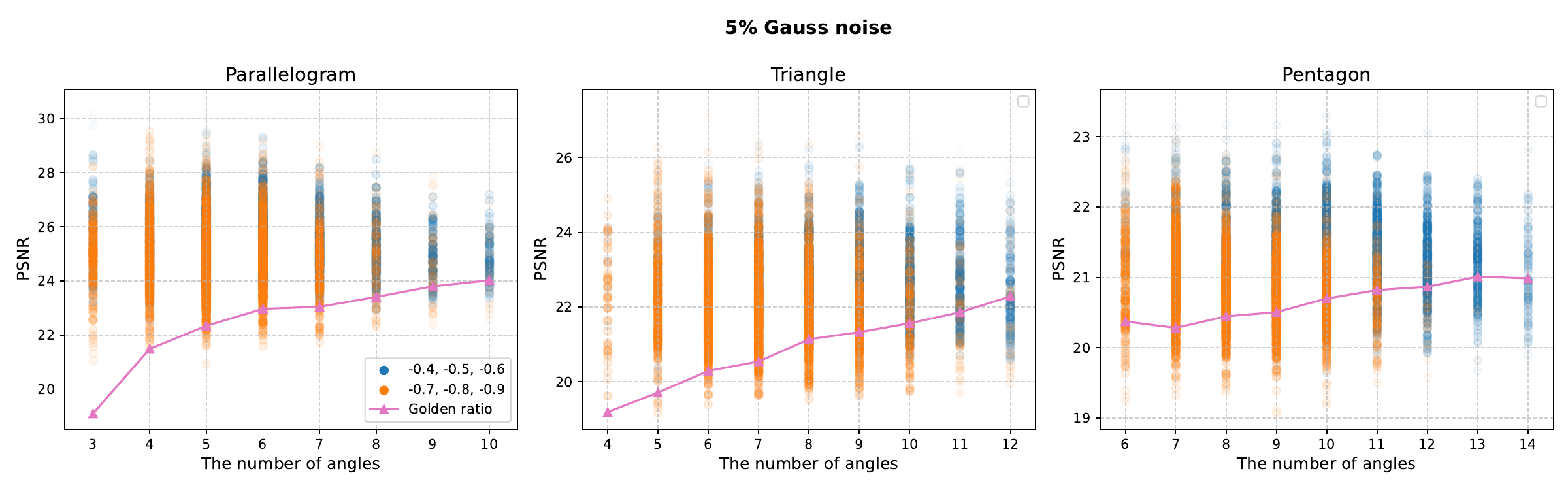}
  \caption{Simulation results are categorized into parallelogram, triangle, and pentagon shapes to illustrate how the number of angles influences PSNR under various rewards and various noise levels. Data points are color-coded by experimental costs---blue ($-0.4, -0.5,  -0.6$) and orange ($-0.7, -0.8, -0.9$). Increased transparency indicates data points that are further from the mean. Triangular markers represent the mean PSNR values obtained from the golden ratio policy at each corresponding angle.}
\label{fig:Validation_comparison}
\end{figure}

Figure (\ref{fig:Validation_comparison}) shows the validation results for experimental costs ranging from \(-0.4\) to \(-0.9\) (in increments of \(0.1\)) under Gaussian noise level of \(5\%\). Results for Gaussian noise levels of \(3\%\) and \(7\%\) are presented in Appendix (\ref{appendix:C}). Each data point represents the relationship between the number of angles and the corresponding PSNR value. For clarity, results are grouped by shape (parallelogram, triangle, and pentagon). The GR policy used the same synthetic data with unseen rotations and the same number of angles chosen by the terminal policy. For each number of angles used by the GR policy, the mean PSNR value was calculated.

Overall, the RL approach with the terminal policy achieved higher PSNR values compared to the GR policy. At lower numbers of angles, the GR policy consistently produced lower mean PSNR values across all shapes, with the difference between the two methods diminishing as the number of angles increased. As shape complexity increased from parallelogram to pentagon, the performance gap between the two policies narrowed. For instance, at lower numbers of angles for the parallelogram, the RL approach yielded data points above the mean value of the GR policy, while for the more complex pentagon shape, the RL policy's advantage was less pronounced. Figure (\ref{fig:Validation_comparison}) uses distinct colors to represent different ranges of experimental costs, with generally lower experimental costs leading to fewer angles. As experimental cost increases, the number of angles decreases across all shapes. In Figure (\ref{fig:Validation_comparison}), the orange points (indicating higher costs) are concentrated at lower numbers of angles, while the blue points (indicating lower costs) are found at higher numbers of angles; however, the distinction is more evident for complex shapes, which require more angles.

Table (\ref{tab:ComaprasionValidation}) summarizes the number of angles selected for the three shapes during validation under different noise levels, with the experimental cost fixed at \(-0.5\). In general, lower noise levels lead to fewer angles being chosen. As shown in the table, each shape requires the fewest angles at \(3\%\) Gaussian noise, whereas \(7\%\) Gaussian noise leads to the largest number of angles. Furthermore, the number of angles increases with the complexity of the geometry: the parallelogram requires the fewest angles, while the pentagon requires the most. The RL policy outperforms the GR policy under every tested condition.

\begin{table}
\centering
\caption{Comparison of the number of angles and PSNR values for triangle and pentagon shapes under different noise levels.}
\label{tab:psnr_comparison}
\begin{tabular}{lllll}
\toprule
        Shape & Noise &    Number of angles & RL (PSNR) & GR (PSNR) \\
\midrule
Parallelogram &    3\% &  6.02 ± 2.26 &    25.96 ± 1.29 &    22.65 ± 2.17 \\
Parallelogram &    5\% &  6.96 ± 2.81 &    25.55 ± 1.10 &    22.92 ± 2.06 \\
Parallelogram &    7\% &  9.41 ± 3.98 &    24.99 ± 0.95 &    23.21 ± 1.91 \\
\midrule
     Triangle &    3\% &  8.54 ± 1.97 &    23.42 ± 1.10 &    21.58 ± 1.71 \\
     Triangle &    5\% &  9.43 ± 2.21 &    22.85 ± 1.09 &    21.44 ± 1.54 \\
     Triangle &    7\% & 11.39 ± 3.09 &    22.05 ± 1.12 &    21.17 ± 1.41 \\
\midrule    
     Pentagon &    3\% & 10.43 ± 1.98 &    22.66 ± 0.67 &    22.21 ± 0.51 \\
     Pentagon &    5\% & 10.99 ± 2.01 &    21.47 ± 0.56 &    20.99 ± 0.51 \\
     Pentagon &    7\% & 12.56 ± 2.25 &    20.01 ± 0.58 &    19.56 ± 0.52 \\
\bottomrule
\label{tab:ComaprasionValidation}
\end{tabular}
\end{table}

\subsection{Test on experimental X-ray CT data with two noise levels}
We explored the gap between simulation and the real world by applying the model trained on the synthetic dataset to the experimental X-ray CT data described in Section 5.1.2. Building on the potential demonstrated in the simulation, we further investigated whether the trained model could adapt to changes in the real scanning environment, particularly variations in noise levels. Two noise levels were considered, based on different emission current settings (600 \(\mu\)A and 100 \(\mu\)A). The model was trained using 5\% Gaussian noise on the projections, while the noise in the experimental X-ray CT typically consists of a mixture of Poisson and Gaussian noise \cite{andriiashen2024x}. It is important to demonstrate whether training under simplified conditions (Gaussian noise model) can produce reasonable results when applied to the more complex conditions of the real world (experimental noise).

\begin{figure}[H]
  \centering
  \includegraphics[width=\linewidth]{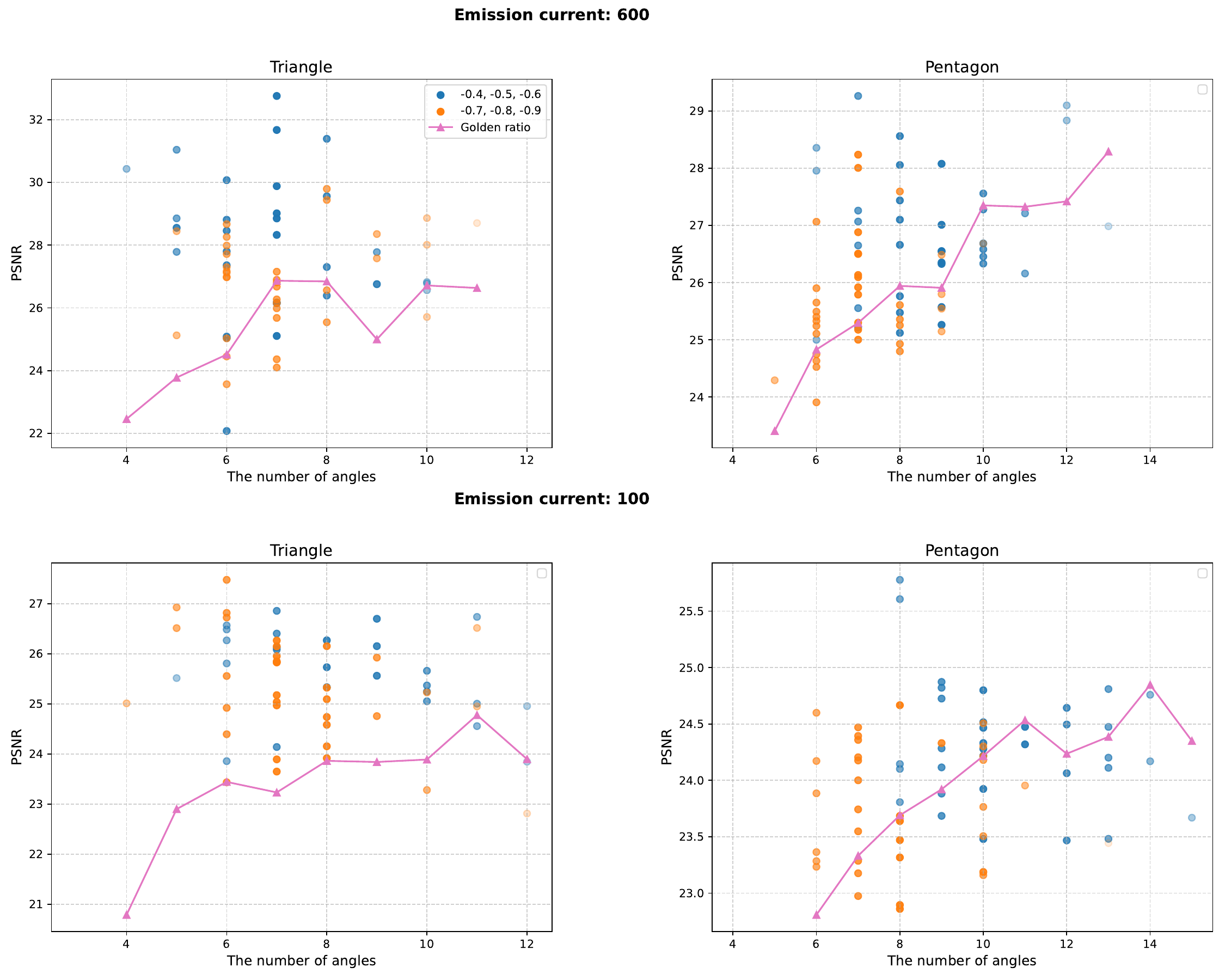}
  \caption{Results on experimental X-ray CT for the triangle and pentagon shapes are shown, illustrating how the number of angles affects PSNR under various experimental costs and noise levels. Data points are color-coded by their experimental costs: blue (\(-0.4, -0.5, -0.6\)) and orange (\(-0.7, -0.8, -0.9\)). Triangular markers represent the mean PSNR values obtained from the GR policy at each corresponding number of angles.
}
\label{fig:Test_comparison}
\end{figure}

Figure (\ref{fig:Test_comparison}) shows results obtained from experimental X-ray CT data using three trained models under varying emission currents and reward settings. To evaluate reconstruction quality, we used the reconstruction obtained with all 180 angles as the ground truth. The GR policy used the same number of angles chosen by the terminal policy. For each number of angles used by the GR policy, the mean PSNR value was calculated. Several consistent trends emerged across the 12 data groups, aligning with the conclusions from the validation. First, pentagon data points generally involved a larger number of selected angles than triangles. Second, noisier data from the lower emission current (100~\(\mu\)A) triggered more angles than data collected at the higher emission current (600~\(\mu\)A). Third, although the number of samples was limited, it was still evident that lower experimental costs (e.g., \(-0.4\), \(-0.5\) and \(-0.6\)) led to more angles, while higher costs (e.g., \(-0.7\), \(-0.8\) and \(-0.9\)) required fewer angles. Finally, the RL policy clearly outperformed the GR policy when fewer angles were selected, though the difference between them diminished as the number of angles increased.

Table (\ref{tab:Test_comparison}) compares the performance of the RL policy and GR policy under an emission current of 600 \(\mu\)A and 100 \(\mu\)A under a reward setting of -0.5. The RL policy consistently outperformed the GR policy for triangles, while it was only comparable to the baselines for pentagons. This outcome, likely influenced by the gap between synthetic and experimental X-ray CT data, is consistent with the training results, where the RL policy did not demonstrate a particularly distinct advantage for pentagon shapes. 

\begin{table}
\centering
\caption{Comparison of the number of selected angles and PSNR values for triangle and pentagon shapes under two different noise levels: The low noise level corresponds to an emission current of 600 \(\mu\)A, and the high noise level to 100 \(\mu\)A.}
\label{tab:psnr_comparison}
\begin{tabular}{lllll}
\toprule
   Shape & Noise &  Number of angles & RL (PSNR) & GR (PSNR) \\
\midrule
Triangle &   Low &  7.92 ± 3.04 &    27.38 ± 2.09 &    25.58 ± 2.10 \\
Triangle &   High & 11.08 ± 2.87 &    25.95 ± 0.72 &    24.47 ± 1.32 \\
\midrule
Pentagon &   Low &  9.17 ± 1.86 &    26.63 ± 1.13 &    26.84 ± 0.87 \\
Pentagon &   High & 10.33 ± 2.05 &    24.50 ± 0.63 &    24.40 ± 0.30 \\
\bottomrule
\end{tabular}
\label{tab:Test_comparison}
\end{table}

\begin{figure}[H]
  \centering
  \includegraphics[width=\linewidth]{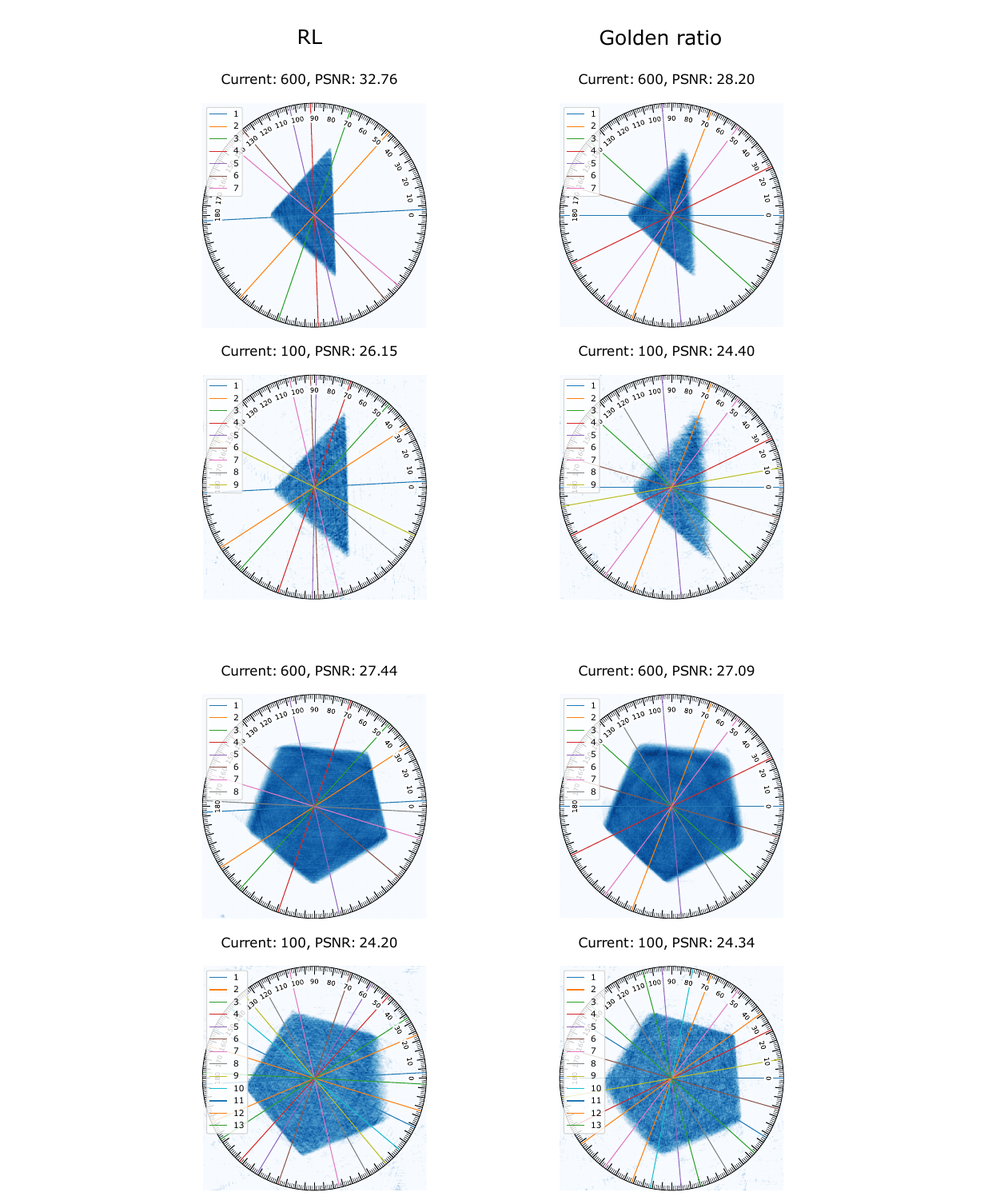}
  \caption{Reconstruction samples on experimental X-ray CT under emission current 100 \(\mu\)A and 600 \(\mu\)A. Colors encode the selection order of projection angles; the legend indicates the corresponding order for each color.}
  \label{fig:reconstruction_comparison}
\end{figure}

Additionally, Figure (\ref{fig:reconstruction_comparison}) illustrates sample reconstructions for the three policies, with the number of angles determined by the RL policy. The results show that the angles selected by the RL policy tend to cluster around the edges. Furthermore, as the noise level increases (from an emission current of 600 \(\mu\)A to 100 \(\mu\)A), the number of selected angles increases, and their distribution becomes broader.

Consequently, the RL policy trained on synthetic data with a simple Gaussian noise model demonstrated reasonable performance on experimental X-ray CT data, effectively handling both optimal stopping and the selection of informative angles.

\section{Discussion}
These results demonstrate that our algorithm effectively balances experimental costs and reconstruction quality by incorporating optimal stopping and adaptively selecting informative angles through deep reinforcement learning. Moreover, the model trained on a synthetic dataset with a simple Gaussian noise model showed satisfactory performance on experimental X-ray CT data, which typically includes more complex noise, such as a mixture of Gaussian and Poisson noise. The algorithm also demonstrated its ability to adapt the number of angles (optimal stopping) to varying noise levels in real-world scenarios while maintaining a consistent level of experimental quality.

Despite these encouraging findings, several areas for improvement remain. First, the gap between simulation and real-world performance could be minimized by incorporating more realistic simulators that account for precise noise models. Second, the training process could be improved by adopting a multi-stage approach, where the model is initially trained on simulated data and subsequently fine-tuned using experimental X-ray CT data. Third, extending the dataset to three-dimensional scenarios would enable the exploration of larger action and state spaces, further enhancing the model's capability. Additionally, task-specific zooming techniques could be integrated for applications such as defect detection. Finally, incorporating learning-based reconstruction algorithms could facilitate the extraction of informative features directly from the reconstruction process, providing a more robust basis for selecting informative angles.

\section{Conclusion}
In this paper, we proposed an approach to simultaneously optimize adaptive informative angle selection and optimal stopping by introducing a terminal policy and jointly computing the policy gradient for both the angle selection and terminal policies. Additionally, we investigated the gap between simulation and real-world scenarios. Our findings demonstrate the feasibility of achieving optimal stopping for sOED based on experimental costs. Furthermore, the trained model from simulation showed promising potential for application to experimental X-ray CT data, highlighting its value for industrial CT applications. This approach paves the way for fully adaptive scanning processes, optimizing both the selection of informative angles and the number of angles required.

\section{Acknowledgement}
This research was co-financed by the European Union H2020-MSCA-ITN-2020 under grant agreement no. 956172 (xCTing).

\section{Data and code availability}

Experimental datasets are available on Zenodo: \url{https://zenodo.org/records/14893740} 

The code for this work is available on GitHub: \url{https://github.com/tianyuan1wang/Optimal_Stopping_RL_CT}

\bibliographystyle{IEEEtran}
\bibliography{main}

\newpage
\appendix
\section{Continuation advantage function}
The following equations show the relationships between the continuation state-value function and the continuation action-value function according to Equation (\ref{eq:ValueStandard}) and Equation (\ref{eq:ReVQ}):
\begin{equation}
    V_{C}^{\pi_{\text{a}}, \pi_{\text{ter}}}(\widehat{\boldsymbol{x}}) = \sum\limits_{\theta}\pi_{\text{a}}(\theta |\widehat{\boldsymbol{x}};\boldsymbol{w}_{a})Q_{C}^{\pi_{\text{a}}, \pi_{\text{ter}}}(\widehat{\boldsymbol{x}}, \theta)
    \label{eq:Relationship1}
\end{equation}

\begin{equation}
    Q_{C}^{\pi_{\text{a}}, \pi_{\text{ter}}}(\widehat{\boldsymbol{x}}, \theta) = -b + \sum\limits_{\widehat{\boldsymbol{x}}^{'}}\pi_{t}(\widehat{\boldsymbol{x}}^{'}|\widehat{\boldsymbol{x}}, \theta) V^{\pi_{\text{a}}, \pi_{\text{ter}}}(\widehat{\boldsymbol{x}}^{'})
    \label{eq:Relationship2}
\end{equation}

Continuation advantage function $A$: It is used to evaluate if the selected action is better than average performance at the current state: 
\label{appendix:A0}
\begin{equation}
\begin{aligned}
    A_{C}^{\pi_{\text{a}}, \pi_{\text{ter}}}(\widehat{\boldsymbol{x}}, \theta) &= \underbrace{Q_{C}^{\pi_{\text{a}}, \pi_{\text{ter}}}(\widehat{\boldsymbol{x}}, \theta)}_{\text{$Q_{C}^{\pi_{\text{a}}, \pi_{\text{ter}}} \rightarrow V_{C}^{\pi_{\text{a}}, \pi_{\text{ter}}}$: Equation (\ref{eq:Relationship2})}} - \underbrace{\sum\limits_{\theta}\pi_{\text{a}}(\theta|\widehat{\boldsymbol{x}};\boldsymbol{w}_{a})Q_{C}^{\pi_{\text{a}}, \pi_{\text{ter}}}(\widehat{\boldsymbol{x}}, \theta)}_{\text{$V_{C}^{\pi_{\text{a}}, \pi_{\text{ter}}} \rightarrow Q_{C}^{\pi_{\text{a}}, \pi_{\text{ter}}}$: Equation (\ref{eq:Relationship1})}} \\
    &=  -b + \underbrace{\sum\limits_{\widehat{\boldsymbol{x}}^{'}}\pi_{t}(\widehat{\boldsymbol{x}}^{'}|\widehat{\boldsymbol{x}}, \theta) V^{\pi_{\text{a}}, \pi_{\text{ter}}}(\widehat{\boldsymbol{x}}^{'})}_{\text{involve terminal policy: Equation (\ref{eq:OS objective})}} - V_{C}^{\pi_{\text{a}}, \pi_{\text{ter}}}(\widehat{\boldsymbol{x}}) \\
    &=  -b + \sum\limits_{\widehat{\boldsymbol{x}}^{'}}\pi_{t}(\widehat{\boldsymbol{x}}^{'}|\widehat{\boldsymbol{x}}, \theta)  \Biggl(\Bigl(1-\pi_{\text{ter}}(d^{'}|\widehat{\boldsymbol{x}}^{'};\boldsymbol{w}_{t})\Bigl)V_{C}^{\pi_{\text{a}}, \pi_{\text{ter}}}(\widehat{\boldsymbol{x}}^{'}) + \pi_{\text{ter}}(d^{'}|\widehat{\boldsymbol{x}}^{'};\boldsymbol{w}_{t}) \mathrm{PSNR}(\widehat{\boldsymbol{x}}^{'}, \Bar{\boldsymbol{x}}) \Biggr) \\
    &- V_{C}^{\pi_{\text{a}}, \pi_{\text{ter}}}(\widehat{\boldsymbol{x}})\\
    & \text{sample $\widehat{\boldsymbol{x}}^{'}$ from  $\pi_{t}$, because we do not know the model $\pi_{t}$} \\
    &\approx  -b +  \Bigl(1-\pi_{\text{ter}}(d^{'}|\widehat{\boldsymbol{x}}^{'};\boldsymbol{w}_{t}) \Bigr)V_{C}^{\pi_{\text{a}}, \pi_{\text{ter}}}(\widehat{\boldsymbol{x}}^{'}) + \pi_{\text{ter}}(d^{'}|\widehat{\boldsymbol{x}}^{'};\boldsymbol{w}_{t}) \mathrm{PSNR}(\widehat{\boldsymbol{x}}^{'}, \Bar{\boldsymbol{x}}) - V_{C}^{\pi_{\text{a}}, \pi_{\text{ter}}}(\widehat{\boldsymbol{x}}).
\end{aligned}
\label{eq:Advantage}
\end{equation}

\section{Policy gradient on \(\boldsymbol{w}_{a}\)}
\label{appendix:A}
\begin{equation}
\begin{aligned}
   \nabla_{\boldsymbol{w}_{a}} J(\boldsymbol{w}_{a}, \boldsymbol{w}_{t})
    &=  \underbrace{\nabla_{\boldsymbol{w}_{a}} 
 V^{\pi_{\text{a}}, \pi_{\text{ter}}}(\widehat{\boldsymbol{x}}_{1})}_{\text{involve terminal policy: Equation (\ref{eq:OS objective})}} \\
    &= \nabla_{\boldsymbol{w}_{a}} \Biggl(\pi_{\text{ter}}(d_{1}|\widehat{\boldsymbol{x}}_{1};\boldsymbol{w}_{t})\mathrm{PSNR}(\widehat{\boldsymbol{x}}_{1}, \Bar{\boldsymbol{x}})+ \Bigl(1-\pi_{\text{ter}}(d_{1}|\widehat{\boldsymbol{x}}_{1};\boldsymbol{w}_{t}) \Bigr)V_{C}^{\pi_{\text{a}}, \pi_{\text{ter}}}(\widehat{\boldsymbol{x}}_{1}) \Biggr) \\
    &=  \Bigl(1 - \pi_{\text{ter}}(d_{1}|\widehat{\boldsymbol{x}}_{1};\boldsymbol{w}_{t}) \Bigr)\nabla_{\boldsymbol{w}_{a}}\underbrace{V_{C}^{\pi_{\text{a}}, \pi_{\text{ter}}}(\widehat{\boldsymbol{x}}_{1})}_{\text{$V_{C}^{\pi_{\text{a}}, \pi_{\text{ter}}} \rightarrow Q_{C}^{\pi_{\text{a}}, \pi_{\text{ter}}}$: Equation (\ref{eq:Relationship1})}} \\
    &=  \Bigl(1 - \pi_{\text{ter}}(d_{1}|\widehat{\boldsymbol{x}}_{1};\boldsymbol{w}_{t}) \Bigr) \Bigl(\sum\limits_{\theta}\nabla_{\boldsymbol{w}_{a}}\pi_{\text{a}}(\theta|\widehat{\boldsymbol{x}}_{1};\boldsymbol{w}_{a})Q_{C}^{\pi_{\text{a}}, \pi_{\text{ter}}}(\widehat{\boldsymbol{x}}_{1}, \theta) \\
    &+ \sum\limits_{\theta} \pi_{\text{a}}(\theta|\widehat{\boldsymbol{x}}_{1};\boldsymbol{w}_{a}) \nabla_{\boldsymbol{w}_{a}}\underbrace{Q_{C}^{\pi_{\text{a}}, \pi_{\text{ter}}}(\widehat{\boldsymbol{x}}_{1}, \theta) \Bigr)}_{\text{$Q_{C}^{\pi_{\text{a}}, \pi_{\text{ter}}} \rightarrow V_{C}^{\pi_{\text{a}}, \pi_{\text{ter}}}$: Equation (\ref{eq:Relationship2})}}\\
    &=  \Bigl(1 - \pi_{\text{ter}}(d_{1}|\widehat{\boldsymbol{x}}_{1};\boldsymbol{w}_{t}) \Bigr) \Bigl(\sum\limits_{\theta}\nabla_{\boldsymbol{w}_{a}}\pi_{\text{a}}(\theta|\widehat{\boldsymbol{x}}_{1};\boldsymbol{w}_{a})Q_{C}^{\pi_{\text{a}}, \pi_{\text{ter}}}(\widehat{\boldsymbol{x}}_{1}, \theta) \\
    &+ \sum\limits_{\theta}\pi_{\text{a}}(\theta|\widehat{\boldsymbol{x}}_{1};\boldsymbol{w}_{a})
    \sum\limits_{\widehat{\boldsymbol{x}}}\pi_{t}(\widehat{\boldsymbol{x}}|\widehat{\boldsymbol{x}}_{1}, \theta_{1}) \underbrace{\nabla_{\boldsymbol{w}_{a}}V^{\pi_{\text{a}}, \pi_{\text{ter}}}(\widehat{\boldsymbol{x}}) \Bigr)}_{\text{unrolling recursive derivation}}
\end{aligned}
\label{eq:GradientW1}
\end{equation}

After the unrolling recursive derivation,
\begin{equation}
\begin{aligned}
   \nabla_{\boldsymbol{w}_{a}} J(\boldsymbol{w}_{a}, \boldsymbol{w}_{t})
    &= \nabla_{\boldsymbol{w}_{a}} V^{\pi_{\text{a}}, \pi_{\text{ter}}}(\widehat{\boldsymbol{x}}_{1}) \\
    &= \sum\limits_{k=1}^{T} \sum\limits_{\theta, \widehat{\boldsymbol{x}}}\pi_{1}^{(k)}(\theta, \widehat{\boldsymbol{x}}, d_{k+1}|\widehat{\boldsymbol{x}}_{k}; \boldsymbol{w}_{a}, \boldsymbol{w}_{t}) \sum\limits_{\theta^{'}}\nabla_{\boldsymbol{w}_{a}}\pi_{\text{a}}(\theta^{'}|\widehat{\boldsymbol{x}}_{k};\boldsymbol{w}_{a})Q_{C}^{\pi_{\text{a}}, \pi_{\text{ter}}}(\widehat{\boldsymbol{x}}, \theta^{'}) \\
    &= \sum\limits_{k=1}^{T} \sum\limits_{\theta, \widehat{\boldsymbol{x}}}\pi_{1}^{(k)}(\theta, \widehat{\boldsymbol{x}}, d_{k+1}|\widehat{\boldsymbol{x}}_{k}; \boldsymbol{w}_{a}, \boldsymbol{w}_{t}) \\&\sum\limits_{\theta^{'}}\pi_{\text{a}}(\theta^{'}|\widehat{\boldsymbol{x}}_{k};\boldsymbol{w}_{a}) \underbrace{\frac{\nabla_{\boldsymbol{w}_{a}}\pi_{\text{a}}(\theta^{'}|\widehat{\boldsymbol{x}}_{k};\boldsymbol{w}_{a})}{\pi_{\text{a}}(\theta^{'}|\widehat{\boldsymbol{x}}_{k};\boldsymbol{w}_{a})}}_{\nabla_{\boldsymbol{w}_{a}}\log\pi_{\text{a}}(\theta^{'}|\widehat{\boldsymbol{x}}_{k};\boldsymbol{w}_{a})}Q_{C}^{\pi_{\text{a}}, \pi_{\text{ter}}}(\widehat{\boldsymbol{x}}, \theta^{'}),
\end{aligned}
\label{eq:GradientW1S_Details}
\end{equation}

where 

\[
\pi_{1}^{(k)} (\theta, \widehat{\boldsymbol{x}}, d_{k+1}|\widehat{\boldsymbol{x}}_{k}; \boldsymbol{w}_{a}, \boldsymbol{w}_{t}) =
\begin{cases}
    \sum\limits_{\theta}\pi_{\text{a}}(\theta|\widehat{\boldsymbol{x}}_{k};\boldsymbol{w}_{a})\sum\limits_{\widehat{\boldsymbol{x}}}\pi_{t}(\widehat{\boldsymbol{x}}|\widehat{\boldsymbol{x}}_{k}, \theta)  \Bigl(1-\pi_{\text{ter}}(d_{k+1}|\widehat{\boldsymbol{x}};\boldsymbol{w}_{t}) \Bigr), & \text{if } k \neq 0, \\
    \Bigl(1-\pi_{\text{ter}}(d_{k+1}|\widehat{\boldsymbol{x}}_{k+1};\boldsymbol{w}_{t})\Bigr), & \text{if } k = 0.
\end{cases}
\]

\section{Policy gradient on \(\boldsymbol{w}_{t}\)}
\label{appendix:B}
\begin{equation}
\begin{aligned}
   \nabla_{\boldsymbol{w}_{t}} J(\boldsymbol{w}_{a}, \boldsymbol{w}_{t})
    &= \underbrace{\nabla_{\boldsymbol{w}_{t}} V^{\pi_{\text{a}}, \pi_{\text{ter}}}(\widehat{\boldsymbol{x}}_{1})}_{\text{involve terminal policy: Equation (\ref{eq:OS objective})}} \\
     &= \nabla_{\boldsymbol{w}_{t}} \Biggl(\pi_{\text{ter}}(d_{1}|\widehat{\boldsymbol{x}}_{1};\boldsymbol{w}_{t})\mathrm{PSNR}(\widehat{\boldsymbol{x}}_{1}, \Bar{\boldsymbol{x}})+ \Bigl(1-\pi_{\text{ter}}(d_{1}|\widehat{\boldsymbol{x}}_{1};\boldsymbol{w}_{t}) \Bigr)V_{C}^{\pi_{\text{a}}, \pi_{\text{ter}}}(\widehat{\boldsymbol{x}}_{1}) \Biggr) \\
     &= \nabla_{\boldsymbol{w}_{t}}\pi_{\text{ter}}(d_{1}|\widehat{\boldsymbol{x}}_{1};\boldsymbol{w}_{t})\mathrm{PSNR}(\widehat{\boldsymbol{x}}_{1}, \Bar{\boldsymbol{x}}) \\
    &-\nabla_{\boldsymbol{w}_{t}}\pi_{\text{ter}}(d_{1}|\widehat{\boldsymbol{x}}_{1};\boldsymbol{w}_{t})V_{C}^{\pi_{\text{a}}, \pi_{\text{ter}}}(\widehat{\boldsymbol{x}}_{1}) +  \Bigl(1 - \pi_{\text{ter}}(d_{1}|\widehat{\boldsymbol{x}}_{1};\boldsymbol{w}_{t}) \Bigr) \nabla_{\boldsymbol{w}_{t}} \underbrace{V_{C}^{\pi_{\text{a}}, \pi_{\text{ter}}}(\widehat{\boldsymbol{x}}_{1})}_{\text{$V_{C}^{\pi_{\text{a}}, \pi_{\text{ter}}} \rightarrow Q_{C}^{\pi_{\text{a}}, \pi_{\text{ter}}}$: Equation (\ref{eq:Relationship1})}}\\ 
    &= \nabla_{\boldsymbol{w}_{t}}\pi_{\text{ter}}(d_{1}|\widehat{\boldsymbol{x}}_{1};\boldsymbol{w}_{t}) \Bigl(\mathrm{PSNR}(\widehat{\boldsymbol{x}}_{1}, \Bar{\boldsymbol{x}})-V_{C}^{\pi_{\text{a}}, \pi_{\text{ter}}}(\widehat{\boldsymbol{x}}_{1}) \Bigr) \\
    &+  \Bigl(1 - \pi_{\text{ter}}(d_{1}|\widehat{\boldsymbol{x}}_{1};\boldsymbol{w}_{t}) \Bigr)\sum\limits_{\theta} \pi_{\text{a}}(\theta|\widehat{\boldsymbol{x}}_{1}; \boldsymbol{w}_{a})\nabla_{\boldsymbol{w}_{t}}\underbrace{Q_{C}^{\pi_{\text{a}}, \pi_{\text{ter}}}(\widehat{\boldsymbol{x}}_{1}, \theta)}_{\text{$Q_{C}^{\pi_{\text{a}}, \pi_{\text{ter}}} \rightarrow V_{C}^{\pi_{\text{a}}, \pi_{\text{ter}}}$: Equation (\ref{eq:Relationship2})}}\\
    &= \nabla_{\boldsymbol{w}_{t}}\pi_{\text{ter}}(d_{1}|\widehat{\boldsymbol{x}}_{1};\boldsymbol{w}_{t}) \Bigl(\mathrm{PSNR}(\widehat{\boldsymbol{x}}_{1}, \Bar{\boldsymbol{x}})-V_{C}^{\pi_{\text{a}}, \pi_{\text{ter}}}(\widehat{\boldsymbol{x}}_{1}) \Bigr) \\
    &+  \Bigl(1 - \pi_{\text{ter}}(d_{1}|\widehat{\boldsymbol{x}}_{1};\boldsymbol{w}_{t}) \Bigr) \sum\limits_{\theta} \pi_{\text{a}}(\theta|\widehat{\boldsymbol{x}}_{1}; \boldsymbol{w}_{a})\sum\limits_{\widehat{\boldsymbol{x}}}\pi_{t}(\widehat{\boldsymbol{x}}|\widehat{\boldsymbol{x}}_{1},\theta) \underbrace{\nabla_{\boldsymbol{w}_{t}} V^{\pi_{\text{a}}, \pi_{\text{ter}}}(\widehat{\boldsymbol{x}})}_{\text{unrolling recursive derivation}} \\
\end{aligned}
\label{eq:GradientW2}
\end{equation}

After the unrolling recursive derivation,
\begin{equation}
\begin{aligned}
   \nabla_{\boldsymbol{w}_{t}} J(\boldsymbol{w}_{a}, \boldsymbol{w}_{t})
    &= \nabla_{\boldsymbol{w}_{t}} V^{\pi_{\text{a}}, \pi_{\text{ter}}}(\widehat{\boldsymbol{x}}) \\
    &= \sum\limits_{k=1}^{T}\sum\limits_{\theta, \widehat{\boldsymbol{x}}}\pi_{2}^{(k)}(\theta, \widehat{\boldsymbol{x}}, d_{k}|\widehat{\boldsymbol{x}}_{k}; \boldsymbol{w}_{a}, \boldsymbol{w}_{t})\nabla_{\boldsymbol{w}_{t}}\pi_{\text{ter}}(d|\widehat{\boldsymbol{x}};\boldsymbol{w}_{t}) \Bigl(\mathrm{PSNR}(\widehat{\boldsymbol{x}}, \Bar{\boldsymbol{x}})-V_{C}^{\pi_{\text{a}}, \pi_{t}}(\widehat{\boldsymbol{x}}) \Bigr),
\end{aligned}
\label{eq:GradientW2S_details}
\end{equation}

where 

\[
\pi_{2}^{(k)} (\theta, \widehat{\boldsymbol{x}}, d_{k}|\widehat{\boldsymbol{x}}_{k}; \boldsymbol{w}_{a}, \boldsymbol{w}_{t}) =
\begin{cases}
   \Bigl(1 - \pi_{\text{ter}}(d_{k}|\widehat{\boldsymbol{x}}_{k};\boldsymbol{w}_{t}) \Bigr)\sum\limits_{\theta}\pi_{\text{a}}(\theta|\widehat{\boldsymbol{x}}_{k};\boldsymbol{w}_{a})\sum\limits_{\widehat{\boldsymbol{x}}}\pi_{t}(\widehat{\boldsymbol{x}}|\widehat{\boldsymbol{x}}_{k},\theta), & \text{if } k \neq 0, \\
    1, & \text{if } k = 0.
\end{cases}
\]

\section{Additional results}
\label{appendix:C}
\begin{figure}[H]
  \centering
  \includegraphics[width=\linewidth]{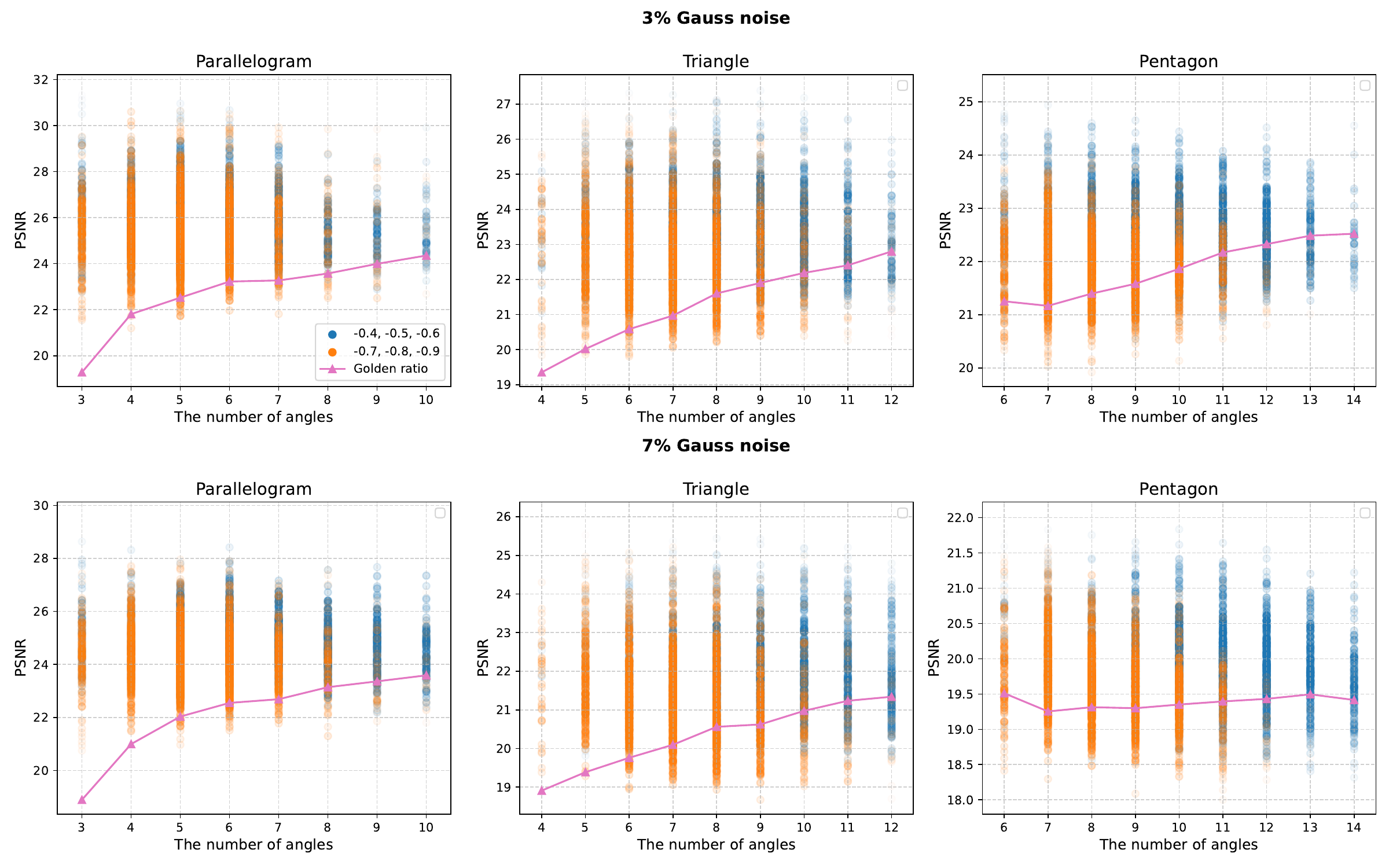}
  \caption{Simulation results are categorized into parallelogram, triangle, and pentagon shapes to illustrate how the number of angles influences PSNR under various rewards and various noise levels. Data points are color-coded by experimental costs---blue ($-0.4, -0.5, -0.6$) and orange ($-0.7, -0.8, -0.9$). Increased transparency indicates data points that are further from the mean. Triangular markers represent the mean PSNR values obtained from the golden ratio policy at each corresponding angle.}
\label{fig:Validation_comparison_appendix}
\end{figure}

\section{Parameters used in datasets}
\label{appendix:D}
\begin{table}[htbp]
\caption{Parameters used in the datasets. The scale is represented by the radius of the circumscribed circles, and the position is described by the center coordinates $(x,y)$ of these circles.} 
\centering
\begin{tabular}{lll}
\toprule
\textbf{Shapes} & \textbf{Scales} & \textbf{Shifts}\\ 
\hline

Parallelogram & Radius: 42 $\sim$ 51 &  Center $(x,y)$: (110,130) $\sim$ (110,130) \\

Triangles & Radius: 56 $\sim$ 89 &  Center $(x,y)$: (110,130) $\sim$ (110,130) \\

Pentagons & Radius: 56 $\sim$ 89 &  Center $(x,y)$: (110,130) $\sim$ (110,130) \\
\bottomrule
\label{tb:DataParameters}
\end{tabular}
\end{table}

\newpage

\section{Architecture of the Actor-Critic network}
\label{appendix:E}
The Actor-Critic model consists of a shared convolutional feature extractor followed by three separate heads for policy, value estimation, and terminal state prediction. The complete architecture is detailed in Table (\ref{tab:actor_critic_architecture}).

\begin{table}[ht]
\centering
\caption{Detailed architecture of the ActorCritic neural network. A shared convolutional encoder extracts features, which are processed by separate heads for policy (actor), value estimation (critic), and terminal prediction.}
\label{tab:actor_critic_architecture}
\begin{tabular}{llllll}
\toprule
\textbf{Layer} & \textbf{Type} & \textbf{Parameters} & \textbf{In Ch.} & \textbf{Out Ch.} & \textbf{Output Size} \\
\midrule
1  & Conv2d     & kernel=3, stride=2, padding=1 & 1   & 12  & $120 \times 120$ \\
2  & GroupNorm  & num\_groups=4                 & 12  & 12  & same \\
3  & LeakyReLU  & negative\_slope=0.2           & 12  & 12  & same \\
4  & MaxPool2d  & kernel=2                      & 12  & 12  & $60 \times 60$ \\
\midrule
5  & Conv2d     & kernel=3, padding=1           & 12  & 24  & same \\
6  & GroupNorm  & num\_groups=4                 & 24  & 24  & same \\
7  & LeakyReLU  & negative\_slope=0.2           & 24  & 24  & same \\
8  & MaxPool2d  & kernel=2                      & 24  & 24  & $30 \times 30$ \\
\midrule
9  & Conv2d     & kernel=3, padding=1           & 24  & 48  & same \\
10 & GroupNorm  & num\_groups=4                 & 48  & 48  & same \\
11 & LeakyReLU  & negative\_slope=0.2           & 48  & 48  & same \\
12 & MaxPool2d  & kernel=4                      & 48  & 48  & $7 \times 7$ \\
\midrule
13 & Flatten    & -                             & -   & -   & $1 \times 2352$ \\
\midrule
\multicolumn{6}{l}{\textbf{Actor Head}} \\
14 & Linear     & in=$1 \times 2352$, out=$1 \times 180$ & - & - & $1 \times 180$  \\
15 & Softmax    & dim=-1                        & -   & -   & $1 \times 180$ \\
\midrule
\multicolumn{6}{l}{\textbf{Critic Head}} \\
16 & Linear     & in=$1 \times 2352$, out=$1 \times 2352$ & - & - & $1 \times 2352$ \\
17 & ReLU       & -                             & -   & -   & same \\
18 & Linear     & in=$1 \times 2352$, out=$1 \times 1$   & -   & -   & $1 \times 1$ \\
\midrule
\multicolumn{6}{l}{\textbf{Terminal Head}} \\
19 & Linear     & in=$1 \times 2352$, out=$1 \times 1$   & -   & -   & $1 \times 1$ \\
20 & Sigmoid    & -                             & -   & -   & $1 \times 1$ \\
\bottomrule
\end{tabular}
\end{table}

\end{document}